\documentclass[default,iicol]{sn-jnl}

\usepackage{caption}
\usepackage{subcaption}
\usepackage{graphicx}%
\usepackage{multirow}%
\usepackage{amsmath,amssymb,amsfonts}%
\usepackage{amsthm}%
\usepackage{mathrsfs}%
\usepackage[title]{appendix}%
\usepackage{xcolor}%
\usepackage{soul}%
\usepackage{textcomp}%
\usepackage{manyfoot}%
\usepackage{booktabs}%
\usepackage{algorithm}%
\usepackage{algorithmicx}%
\usepackage{algpseudocode}%
\usepackage{listings}%
\usepackage{ulem}
\usepackage{url}
\usepackage{listings}
\usepackage{enumitem}
\usepackage{rotating,tabularx}
\usepackage{tikz}
\usepackage{tabularx}
    \newcolumntype{L}{>{\raggedright\arraybackslash}X}
    \newcolumntype{P}[1]{>{\centering\arraybackslash}p{#1}}
\def\checkmark{\tikz\fill[scale=0.4](0,.35) -- (.25,0) -- (1,.7) -- (.25,.15) -- cycle;}

\usepackage{xcolor}
\usepackage{chemfig}

\usepackage{arydshln}

\colorlet{punct}{red!60!black}
\definecolor{background}{HTML}{EEEEEE}
\definecolor{delim}{RGB}{20,105,176}
\colorlet{numb}{magenta!60!black}

\lstdefinelanguage{json}{
    basicstyle=\normalfont\ttfamily\scriptsize,
    numbers=none,
    numberstyle=\scriptsize,
    stepnumber=1,
    numbersep=8pt,
    showstringspaces=false,
    breaklines=true,
    frame=lines,
    backgroundcolor=\color{background},
    literate=
     *{0}{{{\color{numb}0}}}{1}
      {1}{{{\color{numb}1}}}{1}
      {2}{{{\color{numb}2}}}{1}
      {3}{{{\color{numb}3}}}{1}
      {4}{{{\color{numb}4}}}{1}
      {5}{{{\color{numb}5}}}{1}
      {6}{{{\color{numb}6}}}{1}
      {7}{{{\color{numb}7}}}{1}
      {8}{{{\color{numb}8}}}{1}
      {9}{{{\color{numb}9}}}{1}
      {:}{{{\color{punct}{:}}}}{1}
      {,}{{{\color{punct}{,}}}}{1}
      {\{}{{{\color{delim}{\{}}}}{1}
      {\}}{{{\color{delim}{\}}}}}{1}
      {[}{{{\color{delim}{[}}}}{1}
      {]}{{{\color{delim}{]}}}}{1},
}

\makeatletter
\AtBeginDocument{%
  \def\doi#1{\url{https://doi.org/#1}}}
\makeatother

\theoremstyle{thmstyleone}%
\theoremstyle{thmstyletwo}%

\title[]
{ChemScraper: Leveraging PDF Graphics Instructions for Molecular Diagram Parsing}

\author[1]{\fnm{Ayush Kumar} \sur{Shah}}\email{as1211@rit.edu}
\equalcont{These authors contributed equally to this work.}

\author[1]{\fnm{Bryan} \sur{Amador}}\email{ma5339@rit.edu}
\equalcont{These authors contributed equally to this work.}

\author[1]{\fnm{Abhisek} \sur{Dey}}\email{ad4529@rit.edu}

\author[1]{\fnm{Ming} \sur{Creekmore}}\email{mec5765@rit.edu}

\author[2]{\fnm{Blake} \sur{Ocampo}}\email{blakeo2@illinois.edu}
\author[2]{\fnm{Scott} \sur{Denmark}} \email{sdenmark@illinois.edu}

\author[1]{\fnm{Richard} \sur{Zanibbi}}\email{rxzvcs@rit.edu}

\affil[1]{\orgdiv{Document and Pattern Recognition Lab}, \orgname{Rochester Institute of Technology},   \orgaddress{\state{NY}, \country{USA}}}
\affil[2]{\orgdiv{Department of Chemistry}, \orgname{University of Illinois at Urbana-Champaign}, \orgaddress{\state{IL}, \country{USA}}}

\begin{document}

\abstract{
Most molecular diagram parsers recover chemical structure from raster images (e.g., PNGs). However, many PDFs include commands giving explicit locations and shapes for characters, lines, and polygons. We present a new parser that uses these born-digital PDF primitives as input. The parsing model is fast and accurate, and does not require GPUs, Optical Character Recognition (OCR), or vectorization. We use the parser to annotate raster images and then train a new multi-task neural network for recognizing molecules in raster images. 
We evaluate our parsers using SMILES and standard benchmarks, along with a novel evaluation protocol comparing molecular graphs directly that supports automatic error compilation and reveals errors missed by SMILES-based evaluation. 
On the synthetic USPTO benchmark, our born-digital parser obtains a recognition rate of 98.4\% (1\% higher than previous models) and our relatively simple neural parser for raster images obtains a rate of 85\% using less training data than existing neural approaches (thousands vs. millions of molecules). 
}




\maketitle

\section{Introduction}\label{sec1}

We address a pressing need for robust systems to extract molecule drawings
from PDF files. 
Such systems
facilitate data mining applications for chemoinformatics,
multi-modal chemical search, and chemical reaction planning. 

Current molecule structure recognizers generally parse images from
pixel-based raster images, and produce chemical structure descriptions such as Simplified Molecular-Input Line-Entry System strings (SMILES \cite{Weininger1988}) as output. A number of these approaches work well, and some
include modern variations of encoder/decoder models that recognize structure
with high accuracy (see Section \ref{sec:rel_work}).

However, modern documents often use vector images to depict molecules. Vector images encode diagrams as characters, lines, and other graphic
primitives. 
We wish to use PDF drawing instructions directly to produce
fast, accurate methods for indexing molecule images. We were motivated to use PDF instructions by earlier math formula recognition work by Baker et al. using a combination of PDF instructions and image analysis  \cite{BakerSS09}. In our approach, only PDF instructions are used.
In Section \ref{sec:parsing} we describe our improved {\tt SymbolScraper} tool \cite{shah_icdar_2021} that extracts PDF instructions without image processing. 

\begin{figure}[!t]
    \centering
    \includegraphics[width=0.5\textwidth]
    {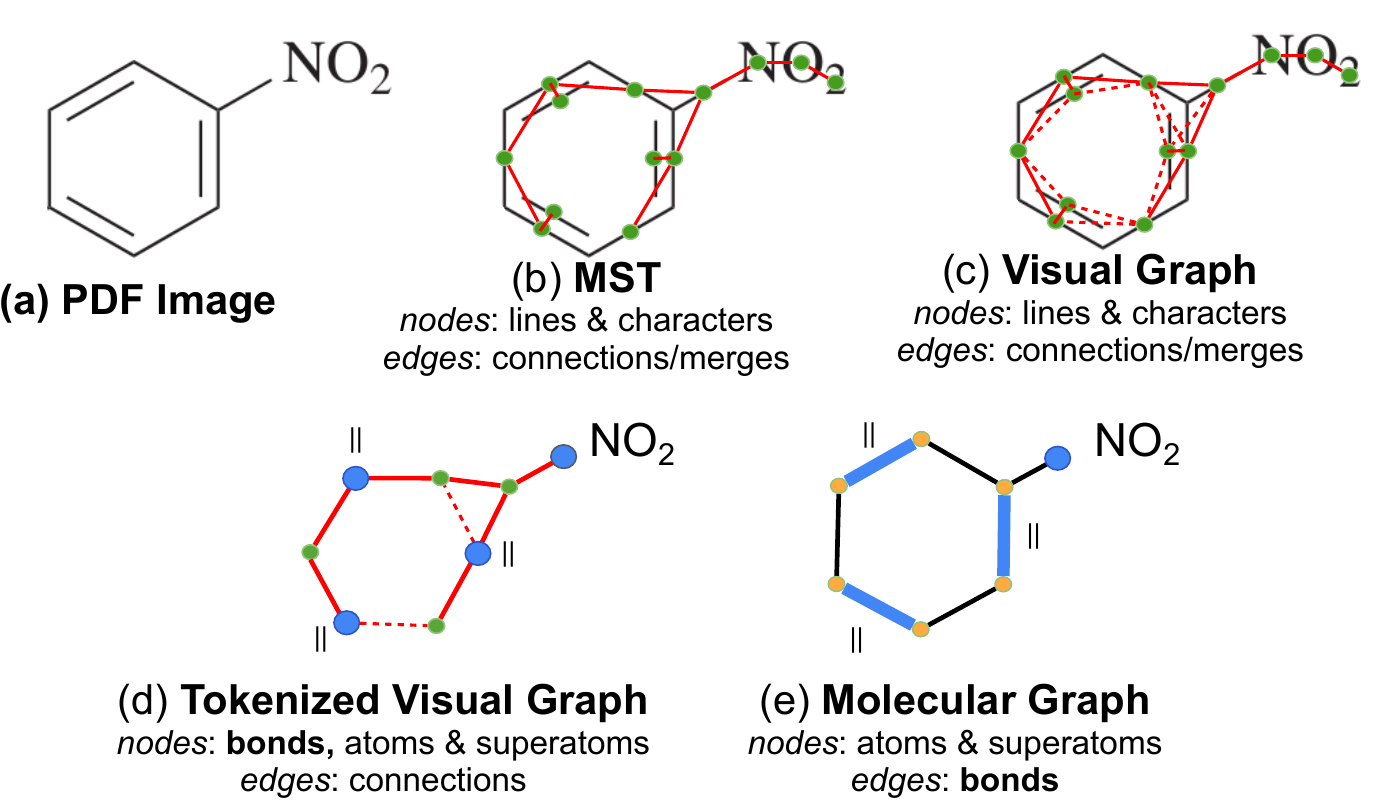}
    \vspace{0.025in}
    \caption{Parsing Nitrobenzene $(C_6H_5NO_2)$ from a PDF image (a). (b) \textbf{Minimum Spanning Tree (MST)} over lines \& characters. 
    (c) \textbf{Visual Graph} with additional edges (dashed lines)
    (d) 
    \textbf{Tokenized Visual Graph} with merged nodes (bonds and named groups). 
(e) \textbf{Molecular Graph}. Blue nodes show 
double bonds and atom/group names in (d) and (e). In (e) orange nodes are `hidden' carbon atoms, and single/double bonds are converted from nodes to edges.}
    \label{fig:mst_to_final}
\end{figure}

In  Section \ref{sec:parsing} we describe the ChemScraper born-digital parser, which is fast and simple in design.\footnote{Publicly available code/tools: \url{https://gitlab.com/dprl/graphics-extraction/-/tree/icdar2024}} As illustrated in Fig.~\ref{fig:mst_to_final}, starting from PDF graphical primitives,  first a Minimum Spanning Tree
(MST) is constructed to identify neighboring primitives. 
Additional edges between primitives are added, and edges to floating objects removed to capture the \textit{visual} structure of the diagram.
Primitives  
are then grouped (i.e., \textit{tokenized})
into molecular entities including atom/superatom names and 
bonds.  Finally, graph transformations convert the tokenized visual graph into a graph 
representing molecular structure.

\begin{figure}[!t]
    \centering
    \includegraphics[width=0.475\textwidth]{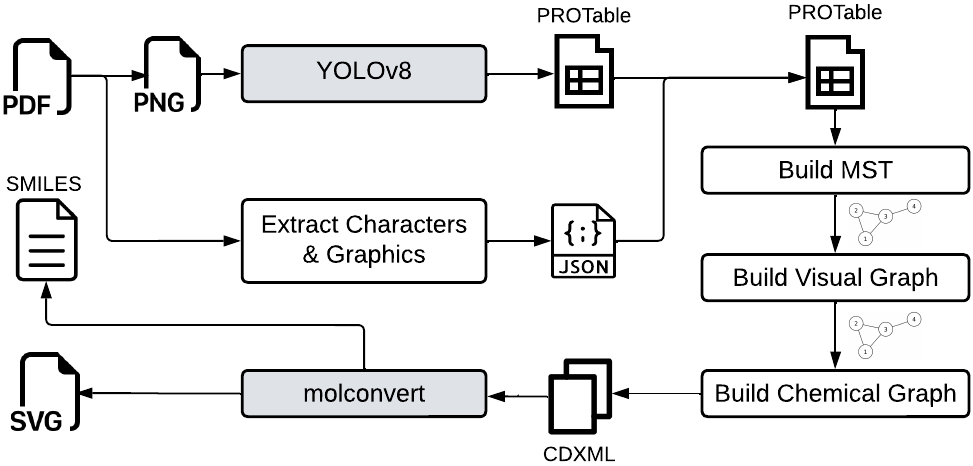}
    \caption{ChemScraper Born-Digital Pipeline. Molecules are detected in PNG page images, but symbols are extracted from PDF instructions. Page-Region-Object tables store bounding boxes and the graphics they contain. Molecules are recognized in three stages, producing CDXML containing the page location, appearance, and chemical structure for each. CDXML can then be converted to chemical structure file formats (e.g., SMILES) or rendered as images (e.g., SVG). }
    \label{fig:pipeline}
\end{figure}

This \textit{born-digital} vector image parser is one component in the online ChemScraper molecule extraction tool\footnote{\url{https://chemscraper.platform.moleculemaker.org/configuration}},
which includes a YOLOv8 \cite{wang_scaled-yolov4_2021} detection module not described in this paper. 
Fig.~\ref{fig:pipeline} provides an overview of the full ChemScraper born-digital extraction pipeline.
The model
locates page regions where molecular diagrams appear, and then parses their 
structure. Recognized molecules are stored in  ChemDraw\footnote{\url{https://revvitysignals.com/products/research/chemdraw}} CDXML files \cite{chemscanner_2019}. CDXML represents both visual and chemical structure in molecular diagrams. The ChemAxon {\tt molconvert} command line tool\footnote{\url{https://docs.chemaxon.com/display/docs/molconvert_index.md}} is used to convert CDXML to vector images (SVG) and SMILES. 
Recognized molecules can then be used for editing, search, and other
applications (e.g., in chemoinformatics). 

We also use the born-digital parser to annotate pixel-based raster images, to
address a shortage of such data. This includes annotations for all graphical
primitives, atoms, and bonds (see Section \ref{sec:data}).  We use this data to
train a new \textit{visual} parser, a novel multi-task neural network for
recognizing molecule diagrams in raster images (see Section \ref{sec:visual}).
The visual parser starts by creating line-shaped contour primitives from a
raster image that over-segment lines and characters. Just as for the
born-digital parser, the  visual parser creates a visual graph providing an
explicit correspondence between an input image and recognized structure, after
which the same tokenization and molecular graph generation steps used for the
born-digital parser are performed. In contrast to recent approaches the neural
network is \textit{segmentation-aware,} and in recurrent runs, input features associated with primitives are updated.
 
In Section \ref{sec:eval}, we evaluate our born-digital and visual parsers with two representations: SMILES and labeled directed graphs. Direct comparison of molecular structure graphs in evaluation is a contribution of this paper: it supports automatic compilation of structural differences. In addition, we report structural differences that are missed in SMILES-based evaluation.

In the next section, we summarize prior work in chemical structure recognition.

\section{Related Work}
\label{sec:rel_work}

We begin by surveying approaches to parsing molecular structure, categorizing them into (1)
rule-based systems, and (2) neural-based systems. For neural-based systems, we further divide these into methods that produce string representations of structure (e.g., SELFIES \cite{Krenn2020}, DeepSMILES \cite{OBoyle2018}, or InChI \cite{Heller2014, Heller2015}) and methods that produce graph representations of structure. 

While our focus is parsing molecular diagrams, we wish to briefly acknowledge
recent work in detecting diagrams. This includes using YOLOv8, an updated
version of Scaled YOLOv4
\cite{wang_scaled-yolov4_2021} with performance and efficiency enhancements.
 In earlier work, Sun et al. \cite{Sun2019b} use 
 a convolutional network, 
addressing scale issues using Spatial Pyramidal Pooling (SPP) \cite{He2015}. Their approach outperformed  
popular detection models of the time, including Faster R-CNN and SSD.

\subsection{Rule-Based Parsers}

The earliest parser for
chemical diagrams in printed documents we know of is a rule-based parser by Ray
et al. from the late 1950's \cite{Ray1957}. This approach first detected atoms in scanned document images, and then connections between atoms were identified
in the regions between atoms. 
Rules based on the number of connections for atoms were used to
determine the type of bonds, which worked well for common compounds.

An important later development was the creation of the Kekul{\'e} system \cite{McDaniel1992}. 
Kekul{\'e} 
adds additional
pre-processing and improved visual detection of bond types over previous methods.
Kekul{\'e} used
thinning and vectorization of raster scans to eliminate variations in bond
lines and characters, and ensured that a consistent set of characters and lines were recovered. Once a connection between a pair of atoms was established, the system
visually detected the bond type instead of using chemical
rules as Ray et al. did. In the same period, CLiDE \cite{Ibison1993} 
added the use of connected component
analysis in disconnected bond groups to identify bond types.
The final adjacency
matrix for structure was created similar to Kekul{\'e}. Another system by Comelli et
al. \cite{Comelli1995} used additional processing to identify 
charges as subscripts or superscripts attached to atoms. 

A still-popular open-source system extending the rules of CLiDE and Kekul{\'e} is OSRA by Filipov et al.
\cite{Filippov2009}.  
OSRA refined processing of raster images generated from born-digital documents, which tend to have clearly rendered text
lines, characters, and graphics. A similar system is MolRec
\cite{Sadawi2012}, which uses horizontal and vertical grouping to detect
connected atoms, their charge, and stereochemical information. 
The more recent CSR system \cite{Bukhari2019} 
also uses rule-based graphical processing
to output SMILES representations for molecules, using the 
\textit{OpenBabel} \cite{oboyle_open_2011} toolkit to generate a valid
connectivity table. 

\subsection{Neural Networks}

{\bf String Output.}
Recent advances in neural networks have proven effective for parsing chemical diagrams. 
For example,
Staker et al. \cite{Staker2019} use an end-to-end model
for extracting molecular diagrams from documents and converting them into SMILES strings. For diagram extraction, they used a U-Net \cite{u_net_2015} to
segment diagrams, which were then passed through an
attention-based encoder network \cite{Vaswani2017AttentionIA} to generate a
SMILES string representing molecular structure from the segmented image.

DECIMER 
\cite{Rajan2020a} also uses an encoder-decoder model for extracting molecular structure from raster images.
In their work they explored using different structure representations, including
SMILES, DeepSMILES, and SELFIES. They found
that SELFIES produced stronger results because of the additional information encoded in comparison with SMILES strings.   

Additional encoder-decoder parsers include IMG2SMI by Campos et al. \cite{Campos2021} which uses a
Resnet-101 \cite{resnet_2015} backbone to extract image features. 
 Li et al. \cite{li_automated_2022} modified a TNT vision
transformer encoder \cite{trans_in_trans} by adding an additional decoder. This use of a vision
transformer was made possible by the BMS (Bristol--Myers--Squibb) dataset \cite{bms_dataset} released by Kaggle, which provided a larger baseline for the conversion of
molecule images to InChI (International Chemical Identifier names).   The training dataset used by Li et al. contained 4 million molecule images. 
Similarly, SwinOCSR by Xu et al. \cite{xuSwinOCSREndtoendOptical2022} used the Swin
transformer to encode image features and another transformer-based decoder to generate
DeepSMILES, and
used a focal loss to address the token imbalance problem in text
representations of molecular diagrams.

{\bf Graph Output.}
String representations of molecular structure lack direct
geometric representation between input objects (e.g., atoms and bonds) and the output strings, and models trained upon them
require extensive training data
\cite{morinMolGrapherGraphbasedVisual2023}. 
In recent years, molecular diagram parsers 
that combine rule-based and
neural-based approaches and generate graph representations have emerged. 
These methods usually employ a
graph decoder or graph construction algorithm.

MolScribe \cite{qian_2023} 
uses a SWIN 
transformer to encode molecular images and a graph decoder consisting of a 
6-layer transformer to jointly predict atoms, bonds, and layouts, yielding
a 2D molecular graph structure. 
They also incorporate rule-based constraints for
chirality (i.e., 3D topology) and algorithms to expand abbreviations. 

MolGrapher \cite{morinMolGrapherGraphbasedVisual2023} is another 
method employing a graph-based output representation. It
utilizes a ResNet-18 backbone to locate atoms, and constructs a supergraph
incorporating all feasible atoms and bonds as nodes, which is then constrained.
Subsequently, a Graph Neural Network (GNN) is applied to the
supergraph, accompanied by external Optical Character Recognition (OCR) for node
classification.
Both these systems utilize multiple data augmentation strategies, including
diverse rendering parameters, such as font, bond width, bond length, and random
transformations of atom groups, bonds, abbreviations, and R-groups (i.e., abbreviations for `rest of molecule') to
bolster model robustness.

Likewise, Yoo et al. \cite{9746088} and OCMR \cite{wangOCMRComprehensiveFramework2023}
produce graph-based outputs directly
from molecular images. 
Yoo et al. \cite{9746088} leverage a ResNet-34 backbone, followed by a
Transformer encoder equipped with
auxiliary atom number and label classifiers. A transformer
graph decoder with self-attention mechanisms is used for bonds. In contrast, Wang
et al. \cite{wangOCMRComprehensiveFramework2023} employ multiple neural network
models for different parsing
steps. These steps include key-point detection, character detection,
abbreviation recognition, atomic group reconstruction, atom and bond prediction.
A graph construction algorithm is subsequently applied to the outputs.

These graph-based methods offer improved
interpretability and robustness, and represent chemical structures naturally.
In particular, atom-level alignment with input images
facilitates easy examination, geometric reasoning, and correction of 
predicted results.

\section{ChemScraper Parsers}

In this paper we present two parsers: one parses molecule diagrams in PDF directly from PDF drawing instructions (vector images), while the other recognizes molecules from raster images (pixel-based). Both parsers use a compiler-style multi-step architecture that (1) identifies input primitives, (2) recovers \textit{visible} diagram structure, 
and  then (3) converts visible structure to  chemical structure information. 

The born-digital parsers' use of Minimum Spanning Trees (MSTs) to recognize molecular diagrams is novel. The detailed PDF graphics information recovered by SymbolScraper is also novel: both as a new data source, and in its application to fast and accurate structure recognition.

To simplify the recognition task, our visual parser operates bottom-up from image region primitives that over-segment lines and characters. The parser is a multi-task, segmentation-aware neural network. The network is run repeatedly until the segmentation (i.e., merging) of primitives remains unchanged. Unlike most recent models, the learning framework utilizes \textit{explicit} segmentation hypotheses, in contrast to `segmentation-free' models generating descriptions of structure without image region correspondences. To support recurrent execution of the network as segmentation changes, we also introduce a novel discrete attention mechanism: images used for classifier input are generated from primitive contours, and are dynamically updated as larger candidate symbols and associated neighborhoods are identified. Similar to other models described above, a ResNet-based convolutional backbone is used for features. However, images of the same size are used for both query and context images, and they are passed separately through the backbone.

Like previous methods, chemical constraints are used to increase accuracy and simplify parser design. Both parsers produce the same visual structure graph representation as illustrated in Fig.~\ref{fig:mst_to_final}(c), and then use the same subsequent steps to tokenize names/bonds and then identify chemical structure.  The regular structure of molecular diagrams motivates using simple visual features, and taking a divide-and-conquer approach to recovering structure. Structure is recovered based on neighboring MST primitives for the born-digital parser, and from small overlapping neighborhoods (windows) in the visual parser.  

An important attribute of ChemScraper output graphs is that they contain both visual and chemical structure information. This allows output graphs to closely match their original appearance in addition to capturing chemical structure. The additional visual information is helpful both for reusing the appearance of molecules within documents, and for visualization and checking of recognition results.

\begin{figure}[!bt]

\parbox{\columnwidth}{
\small
{\bf ~~Input:} Born-Digital PDF Molecule Image 
\begin{enumerate}[font=\small]
\item{\textbf{Extract Symbols from PDF}}\\
Characters and graphical objects (e.g., lines)
\item{\textbf{Build Minimum Spanning Tree (MST)}}\\
Connect neighboring lines, shapes, \& characters
\item{\textbf{MST $\rightarrow$ Visual Graph}} 
\begin{enumerate}[font=\small]
\item Detect negative charges (vs. other lines)
\item Restructure MST\\\textit{(+) add edges}: touching lines (e.g., in rings), adjacent parallel lines and char/line pairs\\\textit{(-) delete edges}: `floating' objects
\item \textbf{Tokenization}\\ 
    $\cdot$ Neighboring characters $\rightarrow$ name nodes \\
    $\cdot$ Neighboring parallel lines $\rightarrow$ bond nodes 
\end{enumerate}
\item { \textbf{Visual Graph $\rightarrow$ Molecular Graph}}\\
{\sc*No tunable parameters}
\begin{enumerate}[font=\small]
\item Convert line intersections into carbons
\item Replace bond nodes by edges
\item Annotate names with subgraphs (e.g., $SO_2$)
\item Generate CDXML 
\end{enumerate}
\end{enumerate}
{\bf Output:} Editable molecular diagram (CDXML)\\
\hrule
}
\caption{Molecule Parsing from PDF Symbols. Symbol information is transformed
into an MST (Fig. \ref{fig:mst_to_final}(b)), a visual structure graph (Fig.
\ref{fig:mst_to_final}(c)), a \textit{tokenized} visual graph (Fig.
\ref{fig:mst_to_final}(d), and finally a molecular structure graph (Fig.
\ref{fig:mst_to_final}(e))}
\label{fig:process}
\end{figure}

\section{Born-Digital Parser}
\label{sec:parsing}

In this section we 
present the ChemScraper born-digital parser for recognizing molecular diagrams directly from vectorized PDF images. 
As seen in Fig.~\ref{fig:process}, our born-digital parser has four stages, including extracting graphics commands using an improved SymbolScraper \cite{shah_icdar_2021}, constructing a Minimum Spanning Tree (MST), rewriting the MST as a visual structure graph, and finally rewriting the visual graph into a molecular structure graph. The final molecular graph replaces line intersections by carbon atoms, and all bond tokens/nodes (e.g., single, double, triple, solid/hashed wedge) are replaced by edges. 

This is a compiler-like recognition architecture, with some similarities to the DRACULAE  mathematical formula recognition system  \cite{zanibbi_2002}. 
Using a compiler-based architecture provides a helpful separation of concerns that allows changes to be implemented and tested across smaller modules.

We provide an overview of the outputs and processing for stages shown in Fig.~\ref{fig:process}. Each stage is then described in more detail in the remainder of this section. The full parsing process has an asymptotic run-time complexity of $O(n^2 \log n)$ for $n$ nodes in the input graph (PDF character/graphics primitives), reflecting the cost of MST construction.

{\bf Stages 1 \& 2: Primitive Graph (MST).}
SymbolScraper recovers primitive symbols from PDF, for which neighboring objects are identified using an MST. Because molecule diagrams represent connections between atoms/groups using line intersections and line/character proximity, MSTs capture many valid connections. However MSTs prune cycles, some primitives must be merged, and some diagrams contain multiple molecules (e.g., parallel lines in bonds and floating ions). 

{\bf Stage 3: (Tokenized) Visual Graph.}
To capture structure missing in the primitive MST, the MST is transformed to provide a two-dimensional syntactic analysis for the visible primitives. This is done by first adding/removing edges to correct MST structure producing a \textit{visual structure graph} (Fig.~\ref{fig:mst_to_final}(c)), 
followed by grouping characters and lines into names and bond types (i.e., tokens) producing a \textit{tokenized visual structure graph}
(Fig.~\ref{fig:mst_to_final}(d)).

{\bf Stage 4: Molecular Graph.}
The final stage is semantic analysis: visual syntax is mapped to represented information/structure, including elements not visible in the diagram. This includes identifying hidden carbon atoms at line intersections, and structures represented only by name. In our system, names are mapped to molecular subgraphs using a dictionary. In Fig.~\ref{fig:mst_to_final}(e), $NO_2$ will be replaced by a subgraph with one nitrogen and two oxygen atoms connected to a hidden carbon. 

The semantic analyzer can also be reused with any parser producing visual graphs in the expected format, and we use this with the visual parser presented later in Section \ref{sec:visual}.

\subsection{Extracting Symbols from PDF}
\texttt{SymbolScraper} is a tool
for extracting characters and shapes  
from vectorized drawing instructions in PDF files, ignoring embedded images \cite{shah_icdar_2021}. This requires
identifying and extracting character shapes (\textit{glyphs}) embedded in font profiles,
as well as instructions for other graphics such as lines and polygons.  Glyphs
and drawing commands define how and where objects
are drawn in a PDF. 
Drawing commands indicate a graphic type (e.g., for font characters, and straight
vs. curved lines).

\begin{figure}[!tb]
   \begin{minipage}{0.48\textwidth}
       \centering
        \includegraphics[width=.35\linewidth]{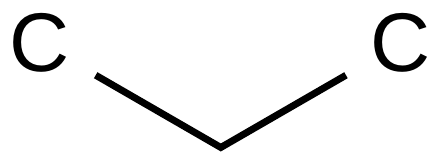}
        \subcaption[l]{Born-Digital PDF for Propane ($C_3H_8$). Note non-visible (implicit) carbon and hydrogen atoms}
   \end{minipage}
    
   \begin{minipage}{0.48\textwidth}
     \centering
     \begin{lstlisting}[label=lst:raw_pdf, language=json]
1 0 0 -1 0 75 cm
45.926 36.102 m
106.832 71.266 l
    \end{lstlisting}
    \subcaption[l]{Instructions for Leftmost Line in PDF Image}
    \end{minipage}    
    \begin{minipage}{0.48\textwidth}
        \centering
        \begin{lstlisting}[label=lst:json, language=json]
{
    "typeFromPDF": "line",
    "graphicObjectID": 0,
    "length": 70.32814383876341,
    "angle": 330.00006986692745,
    "lineWidth": 3.333334,
    "points": [
        {"x": 44.48262170992254,
         "y": 39.73133054974975},
        {"x": 108.27537771024348,
         "y": 2.9006694197326697}
    ]
}
    \end{lstlisting}    
    \subcaption[l]{{\tt SymbolScraper} JSON for Leftmost Line in (a)} 
    \end{minipage}    
    
    \caption{Extracting Symbols from PDF Image (a). 
     (b) \textbf{cm} is a \textit{context matrix}
     defining an affine transformation for subsequent objects. \textbf{m}
     moves the cursor to a point, and \textbf{l} draws a line from the cursor to the
     specified point. (c) Line endpoints, angle, and width are extracted by SymbolScraper.}
     \label{fig:symbols}
\end{figure}

As seen in Fig.~\ref{fig:symbols},
graphic objects in PDF files are defined by instruction sequences. These terminate with an `end-graphic' command (not shown). 
The instructions are in a postfix notation with arguments pushed on a stack before the operations that apply them. Note that coordinates in the JSON output shown in Fig.~\ref{fig:symbols}(c) do not match those in Fig.~\ref{fig:symbols}(b), because the final line endpoints depend upon the line thickness and earlier context matrix. In a larger file, the context matrices are processed cumulatively.

PDF graphics
are defined primarily by instructions for lines, rectangles, and
Bezier curves. We use these as graphical primitives along with their parameters such as
(x,y) points, line widths, whether objects are filled, etc.  
Graphical primitives are converted to 
\textit{line strings} (polylines)\footnote{Java
Topology Suite: \url{https://locationtech.github.io/jts/}}, each of which is a sequence of straight line segments.
We approximate Bezier curves in PDF as straight line segments, using a parameter to limit the maximum distance that a point on the original curve can deviate from the approximated line segments, in points (i.e., 1/72 of an inch).

A small number of rules and additional parameters are used to extract the final input tokens (parameters shown in Table \ref{tab:params}).
Some straight lines are drawn as filled polygons, which are approximated by a line if the two longest lines cover more than a percentage of the polygon perimeter and have their angles within a small tolerance.
Solid wedges (trapezoids) are identified in polygons based on the ratio of long:short side lengths. Positive charges are sometimes drawn with two overlapping lines tested for perpendicularity within an angular tolerance.

The final input tokens produced by SymbolScraper for the born-digital parser are bounding boxes, polygons, or polylines. Each have associated parameters, types, and labels.\footnote{represented using the Python \texttt{Shapely} library}

\begin{table*}[!tb]
\caption{Parameters for PDF Symbol Parsing Stages 
(see Fig. \ref{fig:process}). For visual parsing of raster images (see Section \ref{sec:visual}) only tokenization is applied after creating a structured MST directly.}\label{tab:params}

\centering
\scalebox{0.725}{
\begin{tabular}{ l | P{0.150\linewidth}  P{0.150\linewidth} |  P{0.150\linewidth}
P{0.150\linewidth}  : P{0.150\linewidth} }
\toprule
&
  \multicolumn{2}{c|}{\bf Primitive Graph (MST)} & \multicolumn{3}{c}{\bf 3. MST $\rightarrow$ Visual Graph} \\
~\linebreak \large \sc Parameter (Value) & 1. Extract Symbols & 2. Build MST & (a)\linebreak-ve Charges &(b)\linebreak Restr. MST &(c)\linebreak Tokenization \\

\hline
~ & & & & &\\
\sc PDF GRAPHIC PRIMITIVES & & & & \\
\tt BEZIER\_FLATNESS\_PTS \hfill (0.25) & \checkmark & &  & \\
\tt RECT2LINE\_LONG\_RATIO \hfill (0.85) & \checkmark & & & \\
\tt RECT2LINE\_ANGLE\_TOLERANCE \hfill (5.0) & \checkmark & & &\\
\hline
~ & & & & &\\
\sc ANGLES \& PROXIMITY & & & & &\\ 

\tt ANGLE\_TOLERANCE\_DEGREES \hfill (3.0) &\checkmark & &\checkmark &\checkmark & \checkmark \\ 
\tt CLOSE\_NONPARALLEL\_ALPHA \hfill (1.75) & & & &\checkmark \\ 
\tt CLOSE\_CHAR\_LINE\_ALPHA \hfill (1.5) & & & &\checkmark \\ 
\hline
~ & & & & &\\
\sc SYMBOLS & & & & &\\ 
\tt S-WEDGE\_LENGTHS\_DIFF\_RATIO \hfill (0.7) &\checkmark & & & \\
\tt NEG-CHARGE\_Y\_POSITION \hfill (0.5) & & &\checkmark & \\
\tt NEG-CHARGE\_LENGTH\_TOLERANCE \hfill (0.5) & & &\checkmark & \\

\hline
& & & & & \\
\sc PRUNING EDGES& & & & &\\
\tt ABS\_COS\_CHAR\_PRUNE \hfill (0.1) & & \checkmark & & &\\
\tt CHAR\_LINE\_Z\_TOLERANCE \hfill (1.5) & & \checkmark & & &\checkmark \\
\tt MAX\_ALPHA\_DIST \hfill (2.0) & & & & & \checkmark \\ 
\bottomrule
\end{tabular}
}
\end{table*}

\subsection{Minimum Spanning Tree (MST)} 
\label{sec:mst}

MSTs are widely used for constraints and optimization tasks involving point sets and other geometric object collections in continuous space (i.e., $\mathbb{R}^n$), including agglomerative clustering. For graphics recognition,  MSTs have been used to constrain symbol and spatial relationship types when recognizing handwritten math
formulas, e.g., by Matsakis \cite{Matsakis1999RecognitionOH} and
Eto and Suzuki \cite{EtoS01}.

As can be seen in Fig.~\ref{fig:mst_to_final}, chemical diagrams are even better
suited to MST-based selection of spatial relationships than math formulas. The visual structure of math formulas may have as many as  eight spatial relationship types, while molecule diagrams contain only one spatial relationship ({\tt connected}). Symbols in formulas may be related at a distance, while connections in molecular diagrams are between neighboring symbols. Lines or other graphical objects that need to be combined into symbols (e.g., two parallel lines in a double bond) are also neighboring objects. 

We construct an MST to connect graphical primitives with their nearest neighbor in a chemical  diagram, breaking ties arbitrarily when two or more neighbors are equidistant.
A complete undirected graph over all input PDF primitive pairs is generated first, with edges weighted by distance. 
By default, edge weights are the distance between the closest points on two objects; however, for
line pairs we use their end-points to capture connection distances. This also prevents overlapping lines from having distance 0.

Invalid character connections are prevented by setting distances in our weighted
adjacency matrix to $\infty$ when: (1) The absolute value of the cosine for the
angle between characters falls between [0.1, 0.9], i.e., between [25.8, 84.3]$^\circ$. This prevents (illegal) superscript or subscript character connections. (2) A line-character distance is more than 1.5 standard deviations from the mean line-character distance in the diagram. Pruning parameters are shown in Table \ref{tab:params}.

We use Kruskal's algorithm to extract an MST with $n-1$ edges for $n$ primitives, such that the sum of edge distances is minimal in the pruned adjacency matrix. An example MST over input graphics primitives is shown in Fig.~\ref{fig:mst_to_final}(b). 

\subsection{MST $\rightarrow$ Visual Structure Graph}

While an MST over PDF graphical primitives includes many connections needed to recognize molecular structure,
connections often need to be added or removed. For example, an MST cannot contain cycles, and so we need to insert edges when three or more lines intersect. These and other changes are needed to produce the final graph capturing the visual syntax of a molecular diagram, e.g., as seen in Fig.~\ref{fig:mst_to_final}(d). The steps used for this transformation are presented below; parameters are shown in Table \ref{tab:params}.

{\bf Negative Charges.} We first distinguish negative charges from other lines. Lines are considered negative charges if they are:  (1) roughly horizontal ($0^\circ$), (2) no longer than a fraction of the average line length in the diagram, and (3) right adjacent to a character, with the line's vertical center in the upper half of the character's bounding box.

{\bf Restructure MST.} 
Next we correct connections for `floating' bond lines such as the double bonds in Fig.~\ref{fig:mst_to_final}. These floating lines may not connect with their corresponding parallel line in the MST when another line's endpoint is closer. 
We consider creating an edge between a candidate floating line with degree 1 (one edge) in the MST with another nearby overlapping parallel line if it is within the five nearest neighbors of the line, and the average endpoint distances between the two lines is smaller than for the current neighbor. If so, the line is disconnected from its current neighbor and connected to the closer parallel line.

We then use distance-based clustering to add and remove connections based on MST distances. 

\begin{enumerate}
\item {\it Line Intersections.} Add missing non-parallel line intersections (e.g., for rings and multi-line intersections) where the lines' endpoints are within a ratio of the maximum distance between connected non-parallel lines.

\item {\it Character-Line Connections.} Filter MST char-line connection distances via Z-scores (i.e., standard deviations from the mean) before estimating the maximum char-line connection distance. Add all char-line edges within a ratio of this maximum distance.

\item {\it Split Floating Structures.} Prune edges with a distance larger than a ratio of a maximum distance. The connection type used to determine the maximum distance is selected in the following in order, based on first available distance type in the MST:
(1) char-line distances, (2) parallel line distances, or (3) non-parallel line distances. 
\end{enumerate}

{\bf Tokenization.} There are two steps for merging lines into bonds and characters into atom and group names: (1) merging adjacent characters and parallel lines, and (2) labeling bond types.

{\it Merge Characters and Parallel Lines.}
Characters connected by edges are merged into text tokens, using the location of the nearest character as the connection point for a bond, if present (see Fig.~\ref{fig:mst_to_final}(d)). 
Double bonds, triple bonds, and hashed wedge bonds 
are represented by adjacent parallel lines. Hashed and solid
wedge bonds have a shorter side that begins the bond and a longer side that ends the bond, indicating the bond direction. Solid
wedge bonds are trapezoids, while hashed wedge bonds are drawn as parallel lines of increasing length.
All neighboring parallel line groups in the MST  
are merged, and annotated by the number of lines they contain. For example, in Fig.~\ref{fig:mst_to_final}(d), three pairs of parallel lines representing double bonds will each be merged and annotated with `2'.

{\it Label Bonds in Line Groups/Wedges.}
Annotated line groups can then be labeled as \textit{single}, \textit{double}, or \textit{hashed wedge} bonds by the number of lines they contain (i.e., 1, 2, or $>$3). Three parallel lines are a special case: both triple bonds and hashed wedge bonds may contain 3 parallel lines. We distinguish these by sorting the 3 lines topologically (i.e., top-down, left-to-right), and then determine whether these lines uniformly increase or decrease in size within the sorted list.

For wedge bonds, we need to identify new endpoints on the longest and shortest sides (for solid) or longest and shortest lines (for hashed) and restructure the final visual structure graph accordingly. Bond endpoints are important in the semantic analysis step, which we describe next.

\subsection{Visual $\rightarrow$ Molecular Structure}

In the final stage of the born-digital parser, visual structure is converted to molecular structure, and  chemical information not directly visible in the diagram is added to produce a chemical graph. The chemical graph is then represented in a CDXML file capturing \textit{both} visual and chemical structure. 
Note that this stage uses a deterministic process that involves no tunable parameters.

We first need to define explicit intersection points where line endpoints meet. These intersection points are defined by the midpoint between adjacent endpoints for connected lines in the visual structure graph. `Hidden' carbon atoms are then inserted as nodes at bond line intersections, and at line endpoints without a neighbor. Nodes for bonds in the tokenized visual structure graph are removed, and replaced by edges labeled with the same bond type 
(see Fig. \ref{fig:mst_to_final}(d) and (e)).

{\bf CDXML Generation.}
CDXML is a file format representing molecules and reactions along with related text on a canvas or series of pages. For molecular data, both chemical structure and the appearance of molecules on a 2D canvas are encoded in CDXML files.  
The format was
created for the ChemDraw chemical diagram editor. 

In CDXML tags define molecules, nodes (e.g., atoms, named groups), and bond connections in the diagram, along with annotations for node positions and appearance. We encode the locations of nodes on their associated page, so that the appearance and location of recognized molecules match the original document. Positions are also helpful with accurate conversion to other chemical formats (e.g., SMILES), and to capture spatial information in the chemical structure (e.g., for wedge bonds).

\textit{Annotate Names with Subgraphs}: 
Molecules are often represented more compactly using chemical formulas or other names for substructures. For example,
Fig.~\ref{fig:mst_to_final} shows an abbreviation
node $NO_2$, a nitro group with an external connection available. We use a
manually compiled dictionary of 612 common abbreviations with their associated subgraphs collected from the RDKit Python library\footnote{\url{https://www.rdkit.org}}, ChemDraw, and our own work. For the abbreviation $NO_2$, we insert the full
structure ($*\rightarrow N_1, N_1\rightarrow O_1, N_1\rightarrow O_2$) into the
CDXML as a nested molecule `fragment.' $*$ represents where the structure can be
connected to other structures; $O_1$ and $O_2$ represents two oxygen atoms connected
to the nitrogen $N_1$ through a single and double bond respectively.

\section{Generating Training Data from Visual Graphs}
\label{sec:data}

In designing ChemScraper, we noticed that authors often copy molecular diagrams directly into their documents as raster images, which become embedded in PDFs. To create parsers for raster images with easily interpreted results, we require \textit{explicit} correspondences between image regions and molecular symbols in generated visual structure graphs.   Unfortunately, there is a shortage of training data with direct annotations of raster images.  In addition to fast and accurate recognition, this was the second key motivator for creating our born-digital parser.

While one can create large datasets from SMILES using their rendered raster images, the correspondence between image regions and portions of SMILES strings is absent in such datasets. One can also generate molecular diagram images from MOL files, which include explicit molecular structure (e.g., atoms and their connection by bonds), along with optional 3d spatial positions. However, MOL files were not designed to describe image regions for characters, bonds, or other visual primitives in an image.
For example, MOLs identify spatial locations of atom groups such as
\texttt{$CH_3$}, but do not give the locations for its constituent \texttt{H} and \texttt{3} in an image. 

\begin{figure}[!t]
    \centering
    \includegraphics[width=0.485\textwidth]
    {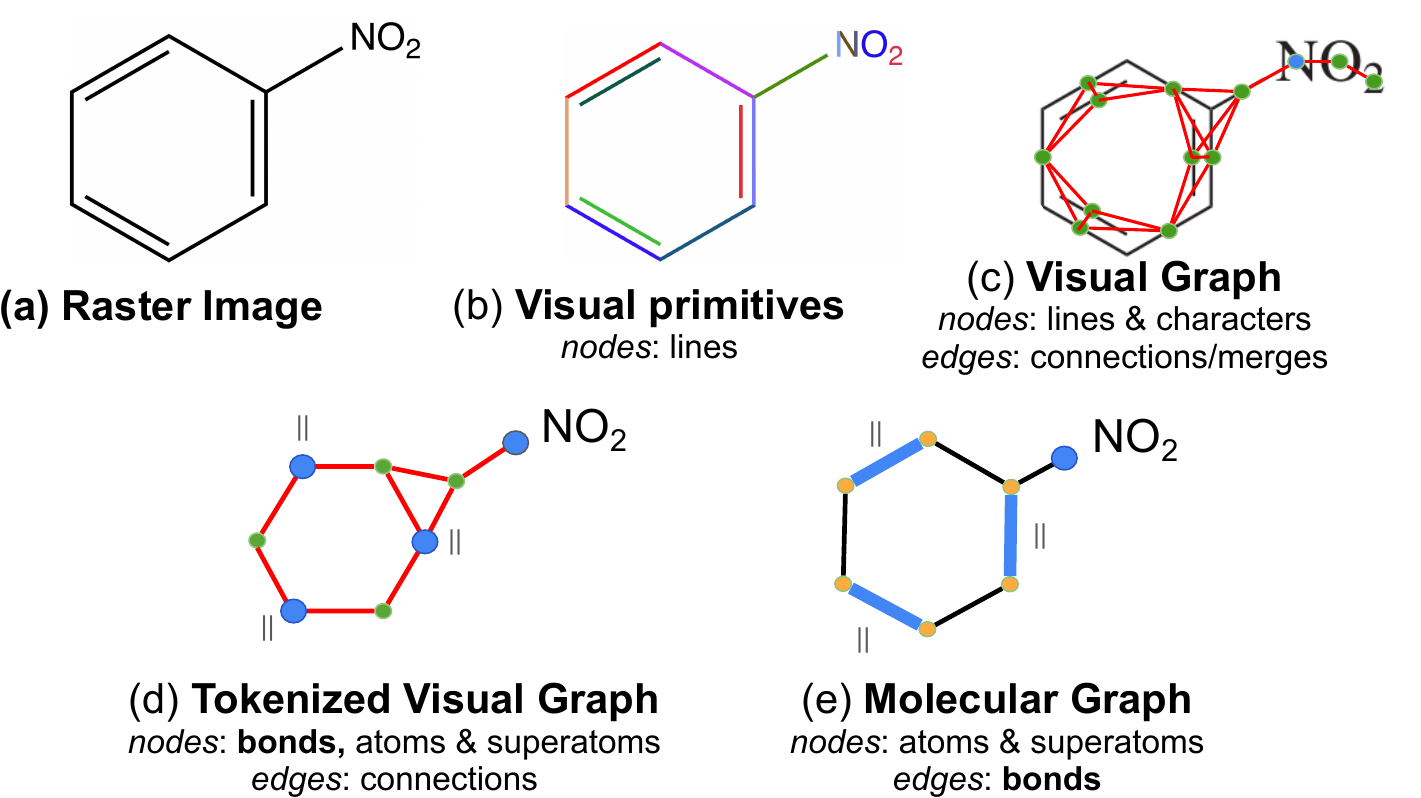}
    \vspace{0.025in}
    \caption{Parsing Nitrobenzene $(C_6H_5NO_2)$ from a raster image (a). (b) \textbf{Visual primitives.} The $N$ is split into 3 lines. 
    (c) 
    \textbf{Visual Graph} extracted from visual parser. 
    (d) 
    \textbf{Tokenized Visual Graph} with merged nodes (bonds and named groups). 
(e) \textbf{Molecular Graph}. Blue nodes show 
the primitives of $N$ merged into a character (c) and 
double bonds and atom/group names in (d) and (e). In (e) orange nodes are `hidden' carbon atoms, and single/double bonds are converted from nodes to edges.}
    \label{fig:visual-parser}
\end{figure}

A new data generation technique is required.
First, we sought a stable visual primitive in pixel-based (raster) molecule images that would avoid merging symbols, and found that we could extract a type of line primitive reliably for this purpose  (see Fig.~\ref{fig:visual-parser}(b)). Given the born-digital parse results for a molecule in PDF, we extract these line primitives from the rasterized PNG for the molecule, and align them with the PDF primitives based on maximum overlap. 

The born-digital visual graphs annotated with line primitives can then be used for training models using the same line primitives as input. For these parsers, the visual primitive extraction replaces the first step of the born-digital parsing pipeline seen in Fig. \ref{fig:pipeline}, where rather than extract characters and lines directly, we may also extract image regions that over-segment (i.e., split) lines and characters. 

{\bf Visual Primitives (Lines).} From a raster image (PNG) for a PDF molecule rendered by the Indigo chemoinformatics toolkit,\footnote{\url{https://github.com/epam/Indigo}} we extract connected component (CC) contours, and convert these to polygons using a simplification algorithm (provided by {\tt Shapely}). These polygons are transformed into a set of skeletal lines using pairs of adjacent parallel lines on the contour boundary. Each pair of parallel lines is replaced by their medial axis (i.e., line between the middle of the parallel lines' endpoints).\footnote{parameters in Table \ref{tab:params} constrain angles and min. overlap}
After the medial axis lines have been identified, pixels in CCs are segmented by assignment to the nearest axis line using a distance transform. 

The resulting `visual' line primitives can be seen in Fig. \ref{fig:visual-parser}(b). Some CC shapes such as curved lines and closed curves are unaltered by the process. The $2$ is unsegmented because after identifying all skeletal lines for CCs in a molecule, to avoid segmenting small CCs, we test whether the average skeletal line length in a CC is less than the average for all skeletal lines. If this average length is smaller than the global average, we do not segment the CC. We also remove skeletal lines within CCs that are smaller than the global average skeletal line length, which avoids over-segmenting lines at dense intersections (e.g., at the connection point between two single bonds and a double bond). We split a long line in a triple or double bond by projecting the floating line onto it, and then testing if the overlap ratio $r$ for the longer line is in the interval of one third to one half, with a margin of $10\%$ ($\frac{1}{3}-\frac{1}{10} \leq r \leq \frac{1}{2}+\frac{1}{10}$). 

For illustration, here we have manually broken the $N$ into three parts; in practice, both characters and lines may be over-segmented.
In Fig.~\ref{fig:visual-parser}(b) there are 15 visual primitives, versus 13 graphical primitives for the original PDF in Figs.~\ref{fig:mst_to_final}(a) and (b). 10 primitives are straight bond lines, and 5 primitives are for the characters in {\tt $NO_2$}.

{\bf Visual Graph Generation.} 
We now annotate raster images using our visual primitives and visual graphs before tokenization (see \ref{fig:mst_to_final}(c)) from our born-digital parser.
We use Indigo to render PDFs from SMILES rather than PNG images as
done in previous methods (e.g., MolScribe \cite{qian_2023}). The born-digital parser is then run on the PDF images, and where the recognized SMILES and original SMILES match (i.e., the result is correct), we use the resulting visual graph as our preliminary ground truth data (e.g., see Fig. \ref{fig:mst_to_final}(c)). 

\begin{figure*}[!tb]
  \centering
  \includegraphics[width=0.96\textwidth]{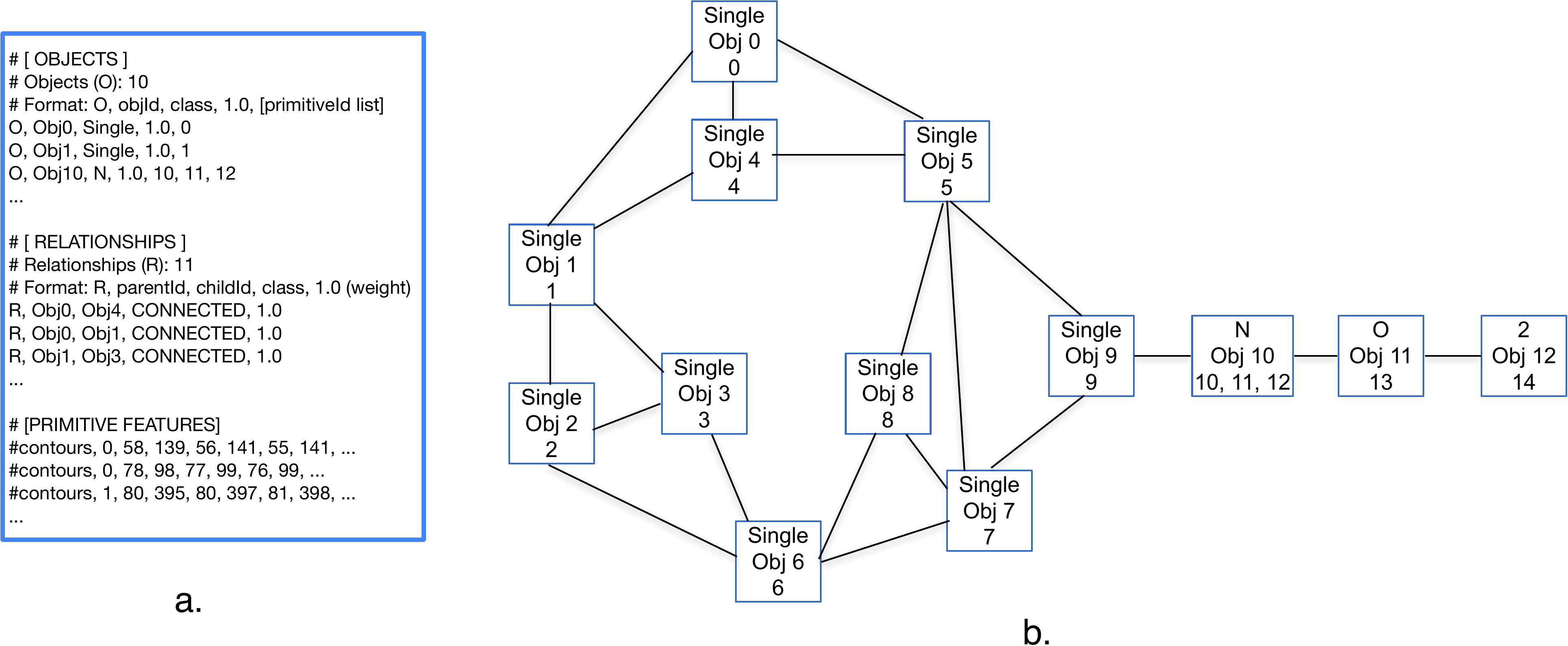}
  \vspace{0.1in}
  \caption{Ground Truth Visual Graph 
  Generated for Fig. \ref{fig:mst_to_final}(c). 
(a) Label graph file with Objects ({\tt O}), Relationships ({\tt R}) and Visual  primitives with 
contour points ({\tt \#contours}). (b) Visualization showing primitive
identifiers, node labels, and edges (all edges labeled as {\tt CONNECTED}). Objects for single bond contain one line primitive each, while the character {\tt N} contains three line primitives. A second file is created using 13 PDF primitives (vs. 15 visual line primitives shown here). }
  \label{fig:gtdata}
\end{figure*}

We next assign visual line primitives to PDF graphical primitives in the born-digital visual graph. PDF images are converted
to 256 DPI PNG images, and we extract visual line primitives as described above.
The assignment of visual primitives to PDF primitives/symbols is determined by maximum overlap. In Figure \ref{fig:visual-parser}(c),
1 line primitive is attached to each line node, 
3 line primitives are attached to $N$, and one primitive is attached to each of the $O$ and $2$. 
Finally, we validate bonds between atoms against a MOL connection table generated from SMILES using Indigo. 

To store visual graphs, we create label graph (Lg) files 
\cite{mouchere_icdar_2013,mouchere_advancing_2016} 
for both PDF primitives and visual line primitives. An example is shown in Fig. \ref{fig:gtdata}(a). Primitives are represented by numeric identifiers and image contours, while typed objects are comprised of one or more primitives (e.g., {\tt Single} bond: one line, character {\tt N}: three lines). 

A label graph file defines structure over declared primitives, using primitive groups (\textit{objects}) and their relationships.
In our label graph files, only {\tt CONNECTED} relationships are explicitly defined, however {\tt MERGE} relationships are  defined implicitly between all primitive pairs in an object. In Fig. \ref{fig:gtdata} {\tt MERGE} edges exist between primitives 10, 11, and 12 for {\tt N} ({\tt Obj10}), and the connection between this character and the {\tt Single} bond {\tt Obj9} is represented by {\tt CONNECTED} edges for (9,10), (9, 11) and (9,12). Similarly, all primitives in an object share a label (e.g., for {\tt Obj10}, primitives 10, 11, and 12 are labeled  {\tt N}).

\section{Visual Parser}
\label{sec:visual}

\begin{figure*}[!t]
    \centering
    \includegraphics[width=0.96\linewidth]
    {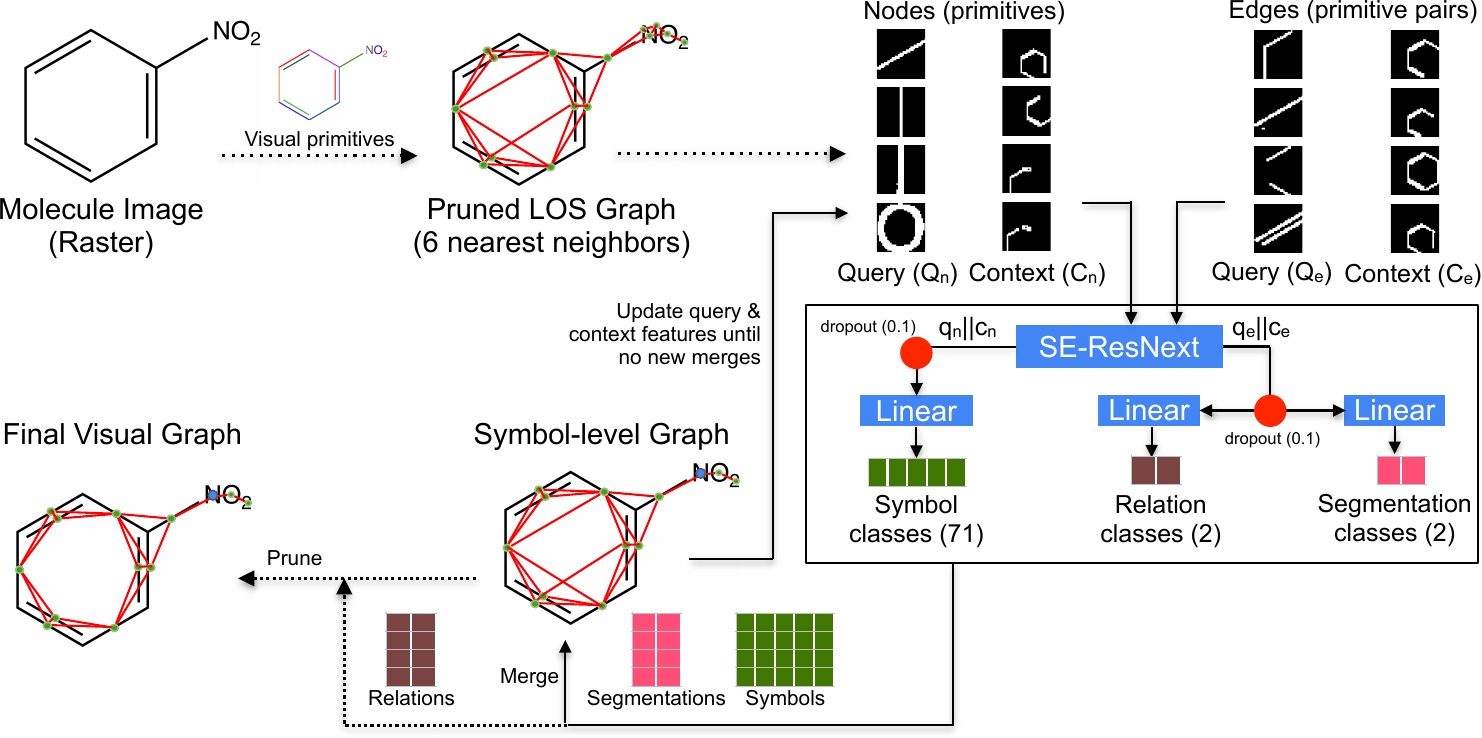}
    \vspace{0.025in}
    \caption{
    Parsing a Raster Image of Nitrobenzene $(C_6H_5NO_2)$. Line contours are extracted as primitives, over which a pruned LOS graph is built.  
    At top-right, four node and four edge queries are shown, at bottom-left their classification tensors (rows: queries, columns: classes). (Q)uery and (C)ontext features enter an SE-ResNext block.
    Two-layer Multi-Layer Perceptrons (MLPs) estimate probabilities for symbol, segmentation ({\small \sc MERGE}), and relationship ({\small \sc CONNECTED}) probabilities. Merges are applied (e.g., for `N'), with symbol/relationship probabilities averaged across primitives.
    The model runs recurrently, updating queries and their contexts until no new merges are found (e.g., two passes for this example). 
 }
    \label{fig:visual-parser-main}
\end{figure*}

In Fig.~\ref{fig:visual-parser-main} we present a  multi-task neural network that parses raster images using the line primitives described in the previous section. The parser produces visual structure graphs, and is trained using our ground truth representation for raster images illustrated in Fig. \ref{fig:gtdata}.  For formulas that contain {\tt MERGE} edges, we use two versions of the input: (1) with no labels, relations, or {\tt MERGE} edges defined (i.e., raw primitive input), and (2) with no labels or relations, but \textit{all} ground-truth {\tt MERGE} edges provided. This allows the model to learn more quickly how to classify symbols and relationships from whole objects rather than their parts.

This parser extends the LGAP model (Line-of-Sight Graph Attention Parser)
\cite{shahLineofSightGraphAttention2023} for parsing
mathematical formulas. The parser creates visual structure graphs by generating
labels for individual primitives and primitive pairs in an input graph, by
classifying individual \textit{queries}.  Compared to the born-digital parser,
the visual parser uses line primitives to replace the first stage of the
pipeline in Fig.~\ref{fig:process}, and the visual parser replaces the second and third stage up to step 3(b) to produce a visual graph ( restructured MST). The remaining tokenization and semantic analysis steps (steps 3(c) and 4) are unchanged.

{\bf Input.}
The parser input is
a Line-of-Sight (LOS) graph over visual line primitives \cite{debergComputationalGeometry2008a, hu_line--sight_2016}, to prune edges between primitives that are `blocked' by a primitive between them. In the LOS graph, edges are defined between primitives where an uninterrupted line may be drawn from the center of one primitive to a point on
the convex hull of the other \cite{mahdavi_lpga_2019}. Connections and merges exist only between nearby primitives in molecular diagrams, as reflected by our use of MSTs in the born-digital parser. Here we prune LOS edges not within the $k=6$ nearest neighbors of a primitive.  There can be at most 4 lines or characters in a bond; we choose 6 neighbors to accommodate over-segmentation in visual primitives.

{\bf Features.}
Visual features are created by drawing line primitive contours directly into $28\times28$  binary images for (1) individual primitives (node queries), (2) primitive pairs (edge queries), and (3) context images containing the $k=6$ nearest neighbors centered around each query (one per node/edge query).
Query and context images are passed separately through a single SE-ResNext backbone producing 32 feature maps per image
\cite{huSqueezeandExcitationNetworks2018,xieAggregatedResidualTransformations2017}.
The first layer of the SE-ResNext encoder is modified, replacing the $7\times7$ convolutional kernel by $3\times3$, using a stride of 1, and {\tt same} padding. We also remove the first maxpool layer because feature images are small.

Feature maps are average pooled in 7 pyramidal regions (image, 3 vertical, 3 horizontal). 
The final query visual features are the pooled convolution responses for a node/edge and its associated context (i.e., $q_n || c_n$ or $q_e || c_e$). For 32 features maps with 7 average-pooled regions, the query and context images produce $2 \times 224 = 448$ features.
We add three positional encodings to query vectors in the form of bounding boxes (BBs) $(x_{min}, y_{min}, x_{max}, y_{max})$ with coordinates normalized to be percentages of width/height: 
\begin{enumerate}
    \item Query BB relative to the formula window
    \item Query BB relative to the context window 
    \item Context window BB relative to the formula 
\end{enumerate}
For edge queries, we use the combined primitive pair position as the query position. Adding these three BBs each query vector contains $448 + (3 \times 4) = 460$ features.  Dropout is applied for regularization (rate of 10\%).

{\bf Classification.}
As seen in Fig.~\ref{fig:visual-parser-main}, node and edge queries are classified using three two-layer multi-layer perceptrons (MLPs):
\begin{enumerate}
    \item Node symbol class (71 class)
    \item Edge primitive merge (2 class)  
    \item Edge primitive connection (2 class) 
\end{enumerate}
For each classification, a hidden linear layer (512 units) is fully connected to the class output layer. 
For 
node queries the 
71 classes include digits, characters, charges (+,-), parentheses, and straight lines. Edge queries are classified twice, once to  identify whether a primitive pair belongs to the same  symbol  ({\tt MERGE}), and then to test whether the primitive are from two connected objects in the diagram ({\tt CONNECTED}). 

{\bf Recurrent Execution.} The parser segments symbols bottom up from input primitives, updating query and context images during recurrent execution. Execution is performed recurrently until edge queries classified as \texttt{MERGE} with probability $> 0.5$ are unchanged from the previous pass (i.e., a fixed point is reached). On a recurrent execution, query images, context images, and positional encodings are all updated for merged primitives. Merges are identified by connected components along {\tt MERGE} edges. 

Note that this is not a conventional recurrent neural network (RNN) where a state vector is updated across executions. Instead, we simply update input features directly as the segmentation changes.
For example, an $N$ broken into three primitives may be merged in the second pass to produce three node queries containing all three primitives. This allows the $N$ to be classified in a single query, rather than in three parts within the first iteration. 
Here the query images are identical for each merged primitive, but note the context image for the primitives will differ because they are centered on the original input primitive associated with each query. This addresses class imbalance by representing multi-primitive symbols multiple times, each with a slightly different context image. 

Recurrent execution stops when no change in \texttt{MERGE} decisions is identified. Edges identified with a probability of being \texttt{CONNECTED} $> 0.5$ are selected; any edges not selected for \texttt{MERGE} or \texttt{CONNECTED} are removed. Symbol and relationship probabilities are then computed by averaging them across primitives in segmented symbols and their connections.

{\bf Training.}  Random over-sampling of node queries is used to balance edge and node queries.   
To balance positive and negative edge examples, we randomly over-sample positive edge examples ({\tt MERGE} and {\tt CONNECTED}), so that each have the same number of positive and negative examples.

Node and edge queries are processed together using a batch size of 64. The sum of cross-entropy losses ($X$) for node and edge queries computed for each batch is
\begin{equation}
\sum_{q_n \in Q_n}  X_S(q_n) ~+~ \sum_{q_e \in Q_e} X_M(q_e) + X_C(q_e)
\end{equation}
where $X_s$, $X_m$ and $X_r$ are the cross entropy loss given the correct target response vectors (1-hot) and softmax distributions for (S)ymbol classification, primitive {\sc (M)erge}, and primitive {\sc (C)onnected} outputs.
For backpropagation, we use an Adam optimizer with learning rate  0.0005, $\beta$ values of (0.9, 0.999), and no weight decay.

\section{Evaluation}
\label{sec:eval}

We next evaluate the accuracy of our parsers.
It is important to remember that the
ChemScraper born-digital parser utilizes PDF information for characters, lines, and other graphical objects that parsers working from raster (pixel) images do not. 
Our analysis includes a graph-based analysis of recognition errors at the level of molecule structure present that provides information missing in standard SMILES-based evaluation methods.

{\textbf{Datasets.}}
For tuning born-digital parser parameters and generating visual parser training data, we use 5000 molecules (46 unique SMILES characters)
extracted from PubChem\footnote{\url{https://pubchem.ncbi.nlm.nih.gov}} prepared by the MolScribe team \cite{qian_2023}. 
For benchmarking, we use three datasets: (1) the USPTO synthetic dataset with 5,179 PNG images generated by the Indigo toolkit from SMILES strings (37 unique SMILES characters) \cite{Rajan2020}, (2) UoB
(5,740 molecule PNG images + SMILES: 33 unique characters \cite{Sadawi2011}), and (3) CLEF (992 molecule PNG images + SMILES: 71 unique characters \cite{Piroi2012}).

The born-digital parser is run on Indigo-rendered PDFs from SMILES ground truth, including for the UoB and CLEF datasets. For the USPTO synthetic set, the rendered PNG and PDF images are essentially identical, but this is not true for the CLEF and UoB data sets where scanned images of molecules were annotated with SMILES; in this case rendering the SMILES using Indigo may produce images in different styles, fonts, and orientations than the scanned molecule images.

Additionally, as described in Section \ref{sec:data}, we generate  annotated visual graph data for training our visual parser that recognizes from raster images.
This comprises 3,416 label graph files from the original pool of 5,000 molecules sourced from PubChem that could be accurately converted into exact SMILES strings. 
Errors include 240 diagrams mis-recognized from valid visual primitives by the born-digital parser, and 1,344 diagrams with errors produced in primitive extraction, alignment, and converting visual graphs to SMILES strings.
This training dataset includes molecules represented by 32 unique symbol classes.
A limitation is that there are
test set symbols missing in this training set.
For the 
USPTO dataset 4 symbols are absent ({\tt 1, a, D, b}), from CLEF 26 symbols are absent (including {\tt *, R, X, 0}), and from the UoB dataset 2 symbols are missing ({\tt :, 0}).

{\textbf{Implementation/Systems.}}
SymbolScraper is built in Java using Apache's {\tt PDFBox} and  the {\tt Java Topology Suite}, while the
ChemScraper born-digital parser is implemented in Python using the
{\tt Shapely} (2d geometry), {\tt networkx} (graphs), {\tt numpy}, and {\tt
mr4mp} (map-reduce) libraries. The ChemScraper born-digial and visual
parsing pipelines are Python-based, along with the visual line primitive extractor.

Born digital parsing
runs were made on a Ubuntu 20.04 server, with a Intel(R) Xeon(R) CPU E5-2667 v4 (3.20 GHz) and 512 GB RAM.
Experiments for the visual parser were run on another Ubuntu 20.04 server with hard drives (HDD), an A40 (48GB) GPU, a 64-core Xeon Gold 6326 (2.9 GHz), and 256 GB RAM.

\subsection{Representations and Metrics}
\label{sec:representations}

We describe the molecule representations and associated metrics used in our evaluation below. 

\paragraph{SMILES Strings: Matches and Similarity}

\label{sec:smiles}
 Simplified Molecular-Input Line-Entry System or SMILES \cite{Weininger1988} represents molecules by the sequence of atoms seen in a traversal of the molecular structure graph. SMILES are compact, and readable for domain experts. 
ChemScraper-generated CDXMLs are first translated to SMILES using ChemAxon's {\tt molconvert} tool. After this, we canonicalize both CDXML and benchmark SMILES to remove differences in their atom order, which can vary for the same molecule. 
SMILES canonicalization is performed
 using the RDKit library via the function {\tt CanonSmiles()}, with {\tt ignore\_chiral=False}.  

SMILES strings are compared by (1) the percentage of exact matches, and (2) the \textit{inverse} of the average
Normalized Levenshtein Distance (NLD).
The \textit{levenshtein distance} is the minimum number of insertions,
  deletions, or substitutions needed to convert one SMILES string to the other \cite{levenshtein}. The distance is
  normalized to $[0,1]$ using the minimum/maximum possible edits based on the SMILES string lengths. The inverse of the average NLD is given by subtracting the average NLD from 1, giving a \textit{similarity} in $[0,1]$, with 1 produced for identical SMILES strings.

\textbf{Limitations.} 
Molecular formulas are naturally represented as
graphs, where atoms and bonds have well-defined relationships and spatial
arrangements. In contrast, SMILES representations are linear character strings \textit{describing} graph structure. These SMILES characters have no direct connection with 
the atoms and bonds present in an input image (i.e., where atoms appear is not represented).

Levenshtein distances for SMILES strings may correspond to multiple operation sequences of the same length. In this case, Levenshtein-based SMILES metrics do not uniquely identify which parts of the input are incorrectly recognized. 
It is thus tempting to instead use graph edit distances over molecule structure graphs directly, with operations that insert/delete/relabel nodes and edges. Unfortunately, this can also result in ambiguous minimal edit sequences, and errors may again not be uniquely identified.

The main issue here is a missing correspondence between input image regions and the nodes/edges in a molecular structure graph representation. If molecular structure graphs include input image locations (e.g., bounding boxes) their nodes may be aligned spatially and then compared using adjacency matrices.  We describe the first application of this approach to chemical structure recognition evaluation next.

\paragraph{Labeled Graphs for Molecular Structure: Label Hamming Distance and Similarity} 
Example molecular structure graphs are shown in Figs.~\ref{fig:mst_to_final}(e) and \ref{fig:visual-parser}(e), which are equivalent.\footnote{\textbf{Note:} The graphs are mostly undirected, but  wedge bonds going `in'/`out' of a page require directed graphs.}  For the ChemScraper parsers, molecular structure graphs produced using born-digital primitives (see Fig.~\ref{fig:mst_to_final}(b)) or visual primitives (see Fig.~\ref{fig:visual-parser}(b)) contain polygons representing the image locations for hidden carbons and atom/group labels. We use these graphs directly for evaluation.

Labeled graphs defined over the \textit{same} nodes with known input locations can be directly compared using their adjacency matrix entries. Recognition errors are easily identified by differing labels in adjacency matrix cells, and located within an input image using the node locations. With a particular bottom-up representation for grouping nodes (i.e., segmentation), errors may be identified even
when node groupings disagree, or nodes are
missing in one or the other graph \cite{lg_icdar2011}. 

Handwritten math formula recognition was evaluated in this manner for the early CROHME competitions, with
ground truth and recognizer outputs defined over the same handwritten strokes \cite{mouchere_advancing_2016}.
The LgEval library\footnote{\url{https://gitlab.com/dprl/lgeval}} was used to compute  metrics and visualize errors  \cite{mouchere_icdar_2013,mouchere_advancing_2016, shah_icdar_2021}.  One can view all errors using the {\tt confHist} tool including missing
nodes and relationships. Repeated errors for nodes, edges, and subgraphs are compiled in histograms that may be explored in HTML pages.

Here we take a slightly different approach. Rather than graphs sharing nodes, corresponding ground truth and output nodes in molecular structure graphs are aligned (i.e., assigned the same identifier) based on spatial overlap in a PDF image. After this alignment, we apply the same adjacency matrix-based evaluation metrics and tools used for CROHME. 

We first assign identifiers to nodes in the ground truth graph, which are atoms or named groups (e.g., $SO_2$) and hidden carbons at line intersections.   
We have adapted MolScribe code to locate atom/group names and hidden carbons in a PDF image for a molecular diagram generated using Indigo. 
Then, parser output graph nodes are given the identifier of the ground truth node  that they have maximum overlap with, breaking ties arbitrarily. Where multiple output nodes overlap one ground truth node, or an output node does not overlap a ground truth node (e.g., missed line intersections produce extra hidden carbons), additional unique identifiers are created.  
Bonds are then defined using labeled edges between nodes using these bond types: (\texttt{single, double, triple, wavy, solid wedge, hashed wedge}).

After alignment, adjacency matrices are used to identify \textit{all}  structural differences from the labels in corresponding cells.  Both rows and columns of adjacency matrices for: (1) ground truth, and (2) parser output, are labeled by the node identifiers obtained during alignment. Node labels are located in diagonal entries (e.g.,  $(n_1, n_1)$) and edge labels are provided in the off-diagonal entries (e.g., $(n_1,n_2)$). For nodes, we compute the percentage of ground truth nodes aligned with an output graph node with the same label  (i.e., (R)ecall), and the percentage of output nodes aligned with an identically labeled ground truth node (i.e., (P)recision). 
We combine Recall and Precision using their harmonic mean $F_1$: $$F_1 = \frac{2RP}{R+P}~.$$ 
We also report the analogous $F_1$ measure for edges (bonds). An output edge is correct if its end nodes and label match ground truth. Finally, we report the percentages of molecules with correct structure (i.e., correct {\tt MERGE} and {\tt CONNECTED} relationships), and with both correct structure and node labels.

\subsection{SMILES-Based Evaluation}

\textbf{Parameter Tuning and Rendering.} 
Each molecule in our 5,000 PubChem molecules for parameter fitting was rendered
with Indigo using 3 randomly selected parameters. The rendering parameters are described below. For benchmarking the born-digital parser, we use the Indigo default rendering parameters. This is done to insure PDF molecules for the born-digital parser have the same appearance as PNG images in the USPTO dataset, which is our primary collection for benchmarking. 

The final parameter values seen earlier in Table \ref{tab:params} are obtained using grid search, with the exception of the {\small \sc PDF GRAPHICS PRIMITIVES} group belonging to {\tt SymbolScraper}. To keep the tuning process manageable, we divided the grid search into 3 stages, one per group in the order given in Table \ref{tab:params}.
Initial default values were identified. After each parameter group's grid search was complete, learned values replaced the default values.
Value ranges and defaults are shown in Table \ref{tab:grid}.

We also tested the effect of the MST pruning parameters discussed in Section \ref{sec:mst}: removing them harms accuracy. For the USPTO dataset removing the absolute cosine angle threshold for characters produces $93.72\%$ SMILES matches, removing the threshold for line-character distances produces $97.06\%$ SMILES, matches and removing both produces $93.20\%$ matches. Including the pruning parameters produces $98.16\%$ exact SMILES matches.

\begin{table}
\caption{Grid Search Parameters. Values tested are shown, with default values in bold.}
\scalebox{0.5}{!}{
\begin{tabular}{l | r}
\toprule
\sc 1. ANGLES \& PROXIMITY \\
\hline
\tt ANGLE\_TOLERANCE\_DEGREES & \{1, 3, \textbf{5}, 10, 15\} \\
\tt CLOSE\_NONPARALLEL\_ALPHA & \{1, 1.25, 1.5, \textbf{1.75}, 2.0\} \\
\tt CLOSE\_CHAR\_LINE\_ALPHA & \{1, 1.25, \textbf{1.5}, 1.75, 2.0\} \\ 
& \\
\sc 2. SYMBOLS \\
\hline
\tt S-WEDGE\_LENGTHS\_DIFF\_RATIO & \{0.70, 0.85, \textbf{0.90}, 0.95\} \\ 
\tt NEG-CHARGE\_Y\_POSITION & \{0, \textbf{0.25}, 0.5\} \\
\tt NEG-CHARGE\_LENGTH\_TOLERANCE & \{0.33, \textbf{0.5}, 0.66\} \\ 
& \\
\sc 3. PRUNING EDGES \\
\hline
\tt ABS\_COS\_CHAR\_PRUNE & \{0.10, \textbf{0.15}, 0.20\} \\
\tt CHAR\_LINE\_Z\_TOLERANCE & \{1.0, \textbf{1.5}, 2.0\} \\
\tt MAX\_ALPHA\_DIST & \{2.0, \textbf{2.5}, 3.0\} \\
\bottomrule
\end{tabular}
}
\label{tab:grid}
\end{table}

\begin{table*}[!tbp]\centering
\caption{Molecular Structure Recognition Benchmarks. Percentages of generated SMILES matching ground truth are shown. For USPTO both PNG and PDF images are rendered using Indigo, but rendered SMILES PDFs may differ from scanned PNGs for CLEF and UoB (indicated by italics).}\label{tab:results}
\scriptsize
\scalebox{0.9}{
\begin{tabular}
{l l  r | r r}
\toprule
& &\sc Synthetic Image &\multicolumn{2}{c}{\sc *Scanned Image} \\\cmidrule{3-5}
\textbf{Models} & &USPTO (5719)  &CLEF-2012 (992) &UoB (5740) \\\midrule
\multirow{3}{*}{Rule-based} &MolVec 0.9.7 &95.40 &83.80 &80.60 \\
&OSRA 2.1 &95.00 &84.60 &78.50 \\
&Imago 2.0 &- &68.20 &63.90 \\ \midrule
\multirow{2}{*}{Neural Network } &Img2Mol &58.90 &48.84 &78.18 \\
&DECIMER &69.60 &62.70 &88.20 \\  \midrule
\multirow{7}{*}{Graph Outputs} &OCMR & - &65.10 &85.50 \\
&SwinOCSR &74.00 &30.00 &44.90 \\
&Image2Graph &- &51.70 &82.90 \\
&MolScribe &\textbf{97.50} &88.90 &87.90 \\
&MolGrapher &- &\textbf{90.50} &\textbf{94.90} \\
\hline
~ & & & \\

\multirow{4}{*}{\textbf{ChemScraper}}& \textbf{Born-Digital Parser (PDF input)}& &  \\
&  ~~(PDF rendering errors) &(15) \textbf{98.16} & \textit{(71) 89.32} & \textit{(0) 94.41} \\
& ~~*Skipping rendering errors &\textbf{98.42}& \textit{96.20} & \textit{94.41} \\
& \textbf{Visual Parser (PNG input)} &85.02 &- &- \\

\bottomrule

\end{tabular}
}
\end{table*}

{\bf Benchmarking: Born-Digital Parser.}
Table \ref{tab:results} compares ChemScraper and existing molecule parsing models. For the USPTO dataset, we see that the born-digital parser obtains the highest rates. Note that the `rendering failure' for USPTO applies to all systems, because the SMILES for these 15 molecules are missing in the collection itself. Given this, the born-digital parser working from PDFs outperforms the neural models working from raster images by nearly 1\%, and rule-based system working from raster images by roughly 3\%. The strong performance of the born-digital parser is because of the additional information available from PDF instructions, and the robust design of the born-digital parser.  

The model also obtains competitive rates for CLEF and UoB, but note that this is for Indigo-rendered SMILES, and not the provided PNGs because PDF images are not provided in these collections. 

In terms of execution time, running the born-digital parser on the USPTO-Indigo dataset (5,719 molecules) with a single process took 28.01 mins (293.39 ms/formula), i.e., 3.4 molecules/sec, with a peak CPU memory use of 230 MB. With multiple processes (32) the total time is reduced to 1.81 mins (19.04 ms/formula), i.e., 52.5 molecules/sec.
Performance benchmarks from Rajan et al. \cite{Rajan2020} show that on a Linux workstation with Ubuntu 20.04 LTS, two Intel Xeon Silver 4114 CPUs and 64 GB of RAM, processing the USPTO-Indigo dataset took 28.65 minutes for MolVec 0.9.7, and 145.04 minutes for OSRA 2.1. Thus, on comparable systems, our born-digital parser operates at  similar or faster speeds compared to other rule-based methods.

\begin{figure}[!tb]
     \centering
     \begin{subfigure}[b]{0.2\textwidth}
         \centering         
         \includegraphics[width=\textwidth]{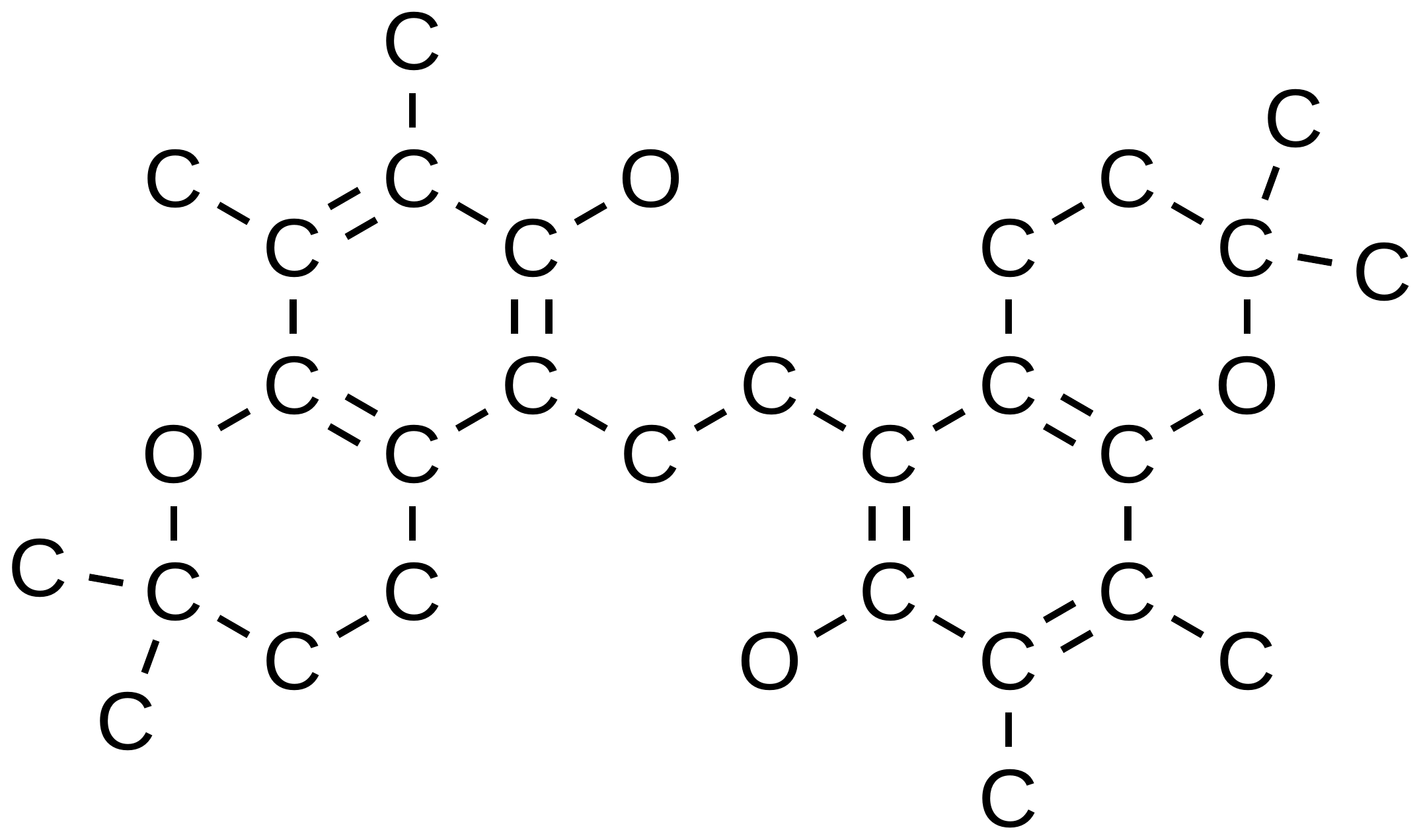}         
         \caption{(\texttt{all}, \texttt{F}, \texttt{1.5})}
         \label{fig:all_False_1.5}
     \end{subfigure}
     \hfill
     \begin{subfigure}[b]{0.2\textwidth}
         \centering
         \includegraphics[width=\textwidth]{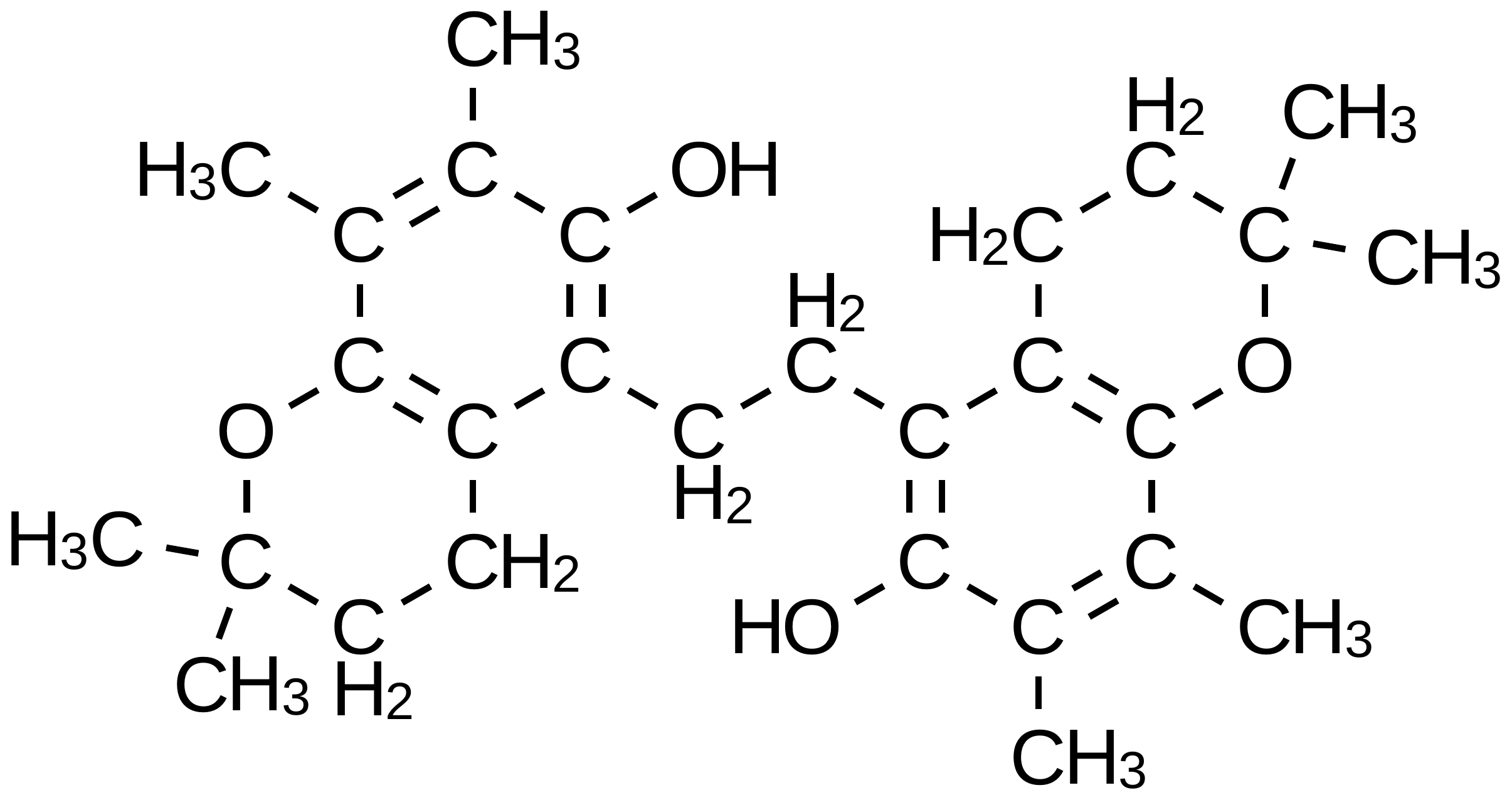}
         \caption{(\texttt{all}, \texttt{T}, \texttt{1.5})}
         \label{fig:all_True_1.5}
     \end{subfigure}
     \hfill
     \begin{subfigure}[b]{0.2\textwidth}
         \centering
         \includegraphics[width=\textwidth]{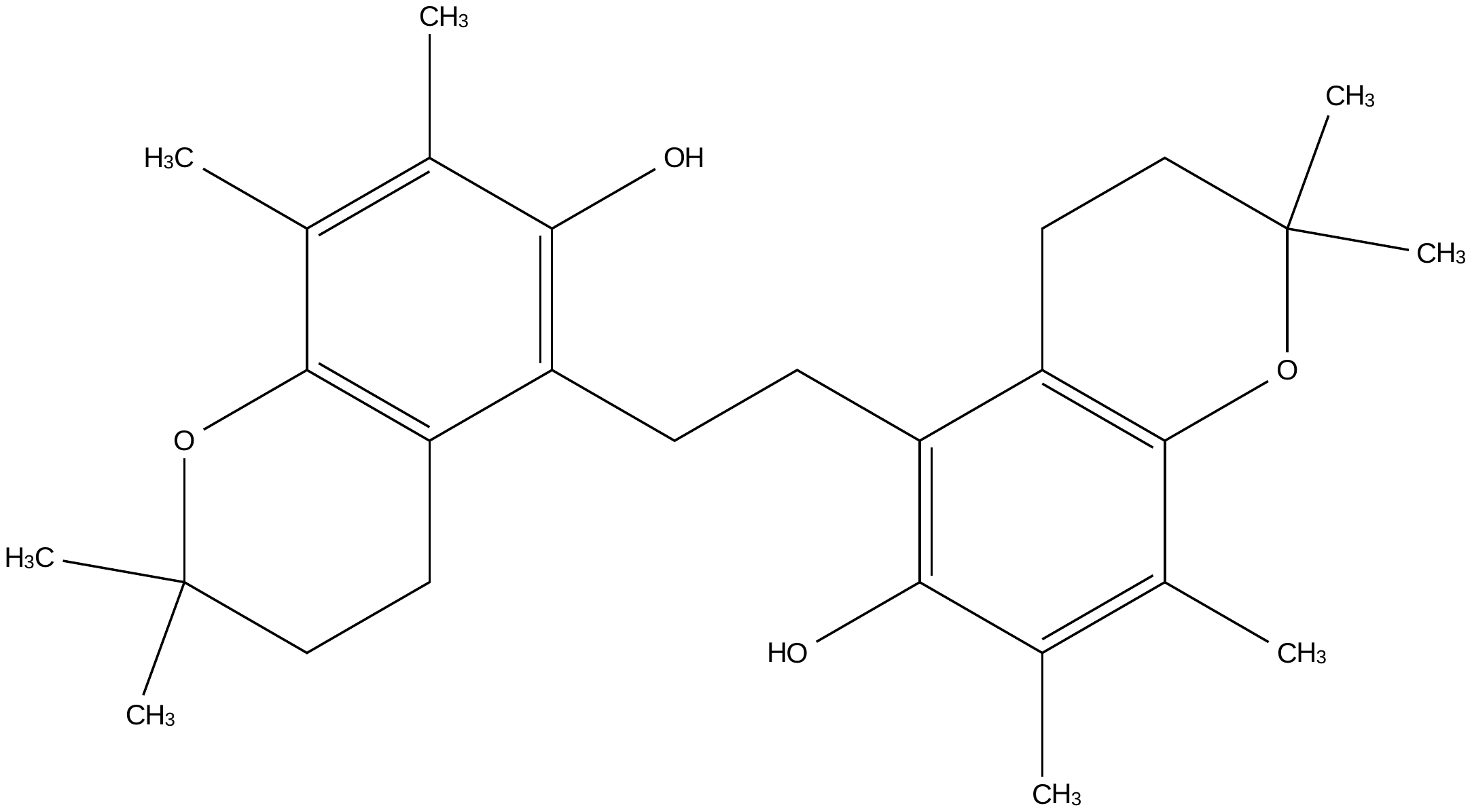}          
         \caption{(\texttt{term-h}, \texttt{T}, \texttt{0.5})}
         \label{fig:terminal-hetero_True_0.5}
     \end{subfigure}
     \hfill
     \begin{subfigure}[b]{0.2\textwidth}
         \centering
         \includegraphics[width=\textwidth]{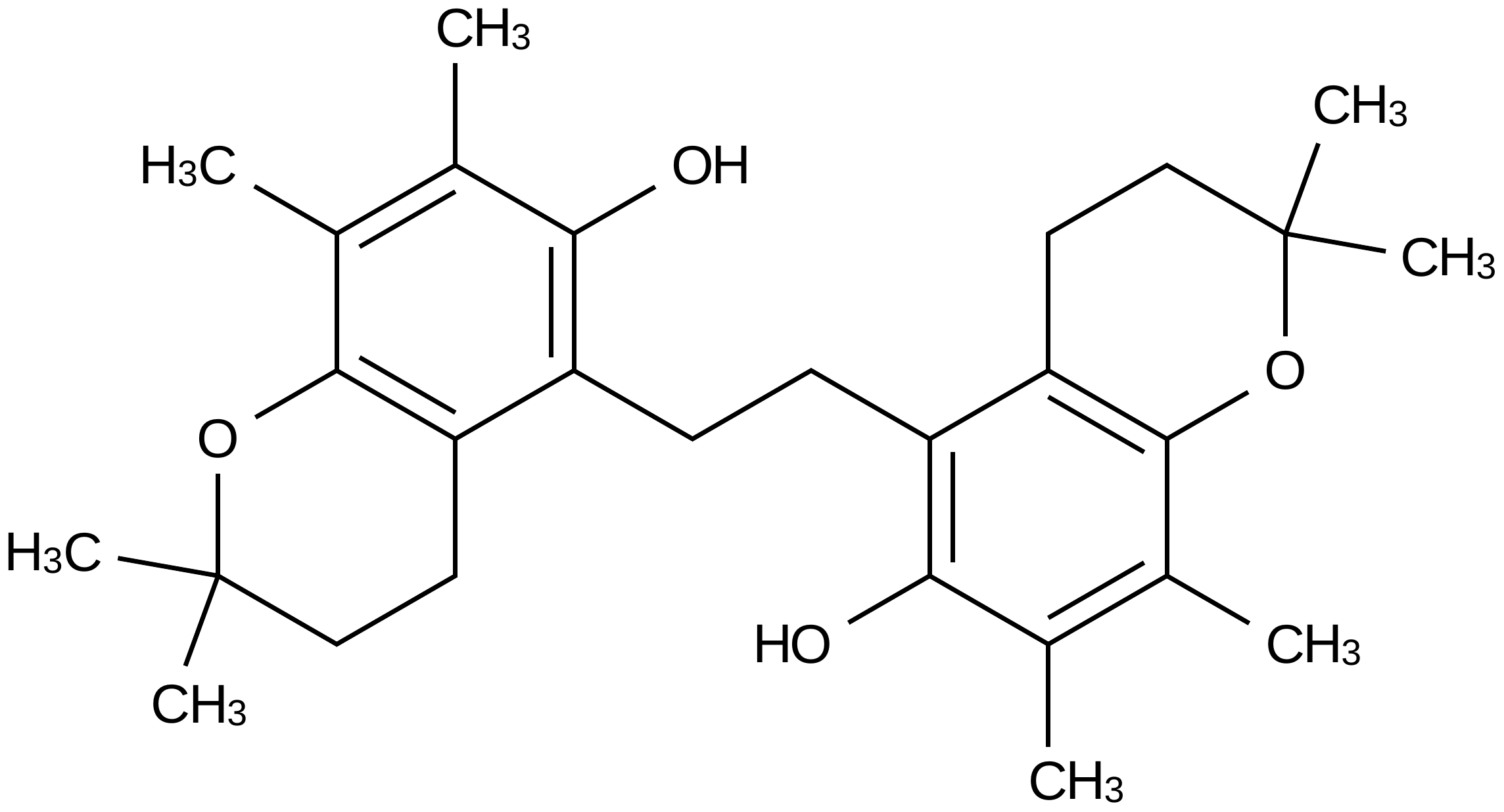}     
         \caption{~(\texttt{term-h}, \texttt{T}, \texttt{1.0})}
         \label{fig:terminal-hetero_True_1.0}
     \end{subfigure}
        \caption{Rendering a molecule with different parameters (Indigo toolkit). Each of (a)-(d) indicate the label mode, whether implicit hydrogens are shown, and the relative thickness. Parameters in (d) are the defaults. The born-digital parser recognizes all four versions correctly.}
        \label{fig:rendering_params}
\end{figure}

{\bf Rendering: Sensitivity Analysis.} To check the robustness of the born-digital parser, we used the rendering parameters of Indigo to perform a sensitivity analysis. We tested three rendering parameters visualized in Fig. \ref{fig:rendering_params}. Parameters/values considered are:
\begin{enumerate}
    \item \texttt{relative-thickness}: Boldness of graphic and text objects. Values considered: \{0.5, 1, 1.5\}. The default is 1.
    \item \texttt{render-implicit-hydrogens-visible}: Whether to show implicit hydrogens. Default is True.
    \item \texttt{render-label-mode}: Which atom labels to show: \{hetero, terminal-hetero, all\}. \textit{all} shows all atoms. There is a \textit{none} option we omit because it leads to ambiguous molecules. Default is \textit{terminal-hetero}.
\end{enumerate}
This produces 18 parameter combinations for rendering. We evaluated our parser with each of them for the  USPTO Indigo dataset, using SMILES matches and inverse normalized levinshtein distances for evaluation.

\begin{figure}[!tb]
    \centering
    \includegraphics[width=0.50\textwidth]{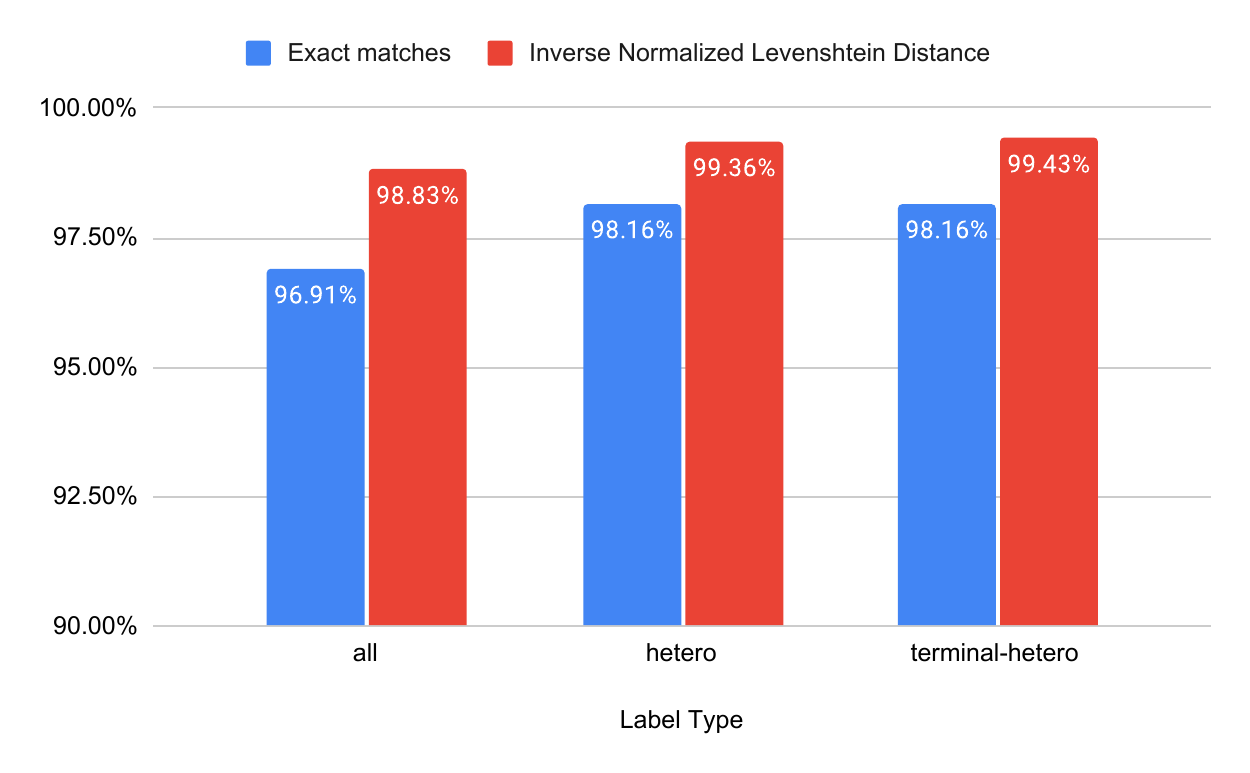}
    \caption{Sensitivity of Born-Digital Parser to Label Rendering Parameter. SMILES-based evaluation is used. Other parameters have default values, with \texttt{render-implicit-hydrogens-visible} as True and \texttt{render-relative-thickness} to 1.}
    \label{fig:label_plot}
\end{figure}

\begin{figure}[!tb]
    \centering
    \includegraphics[width=0.50\textwidth]{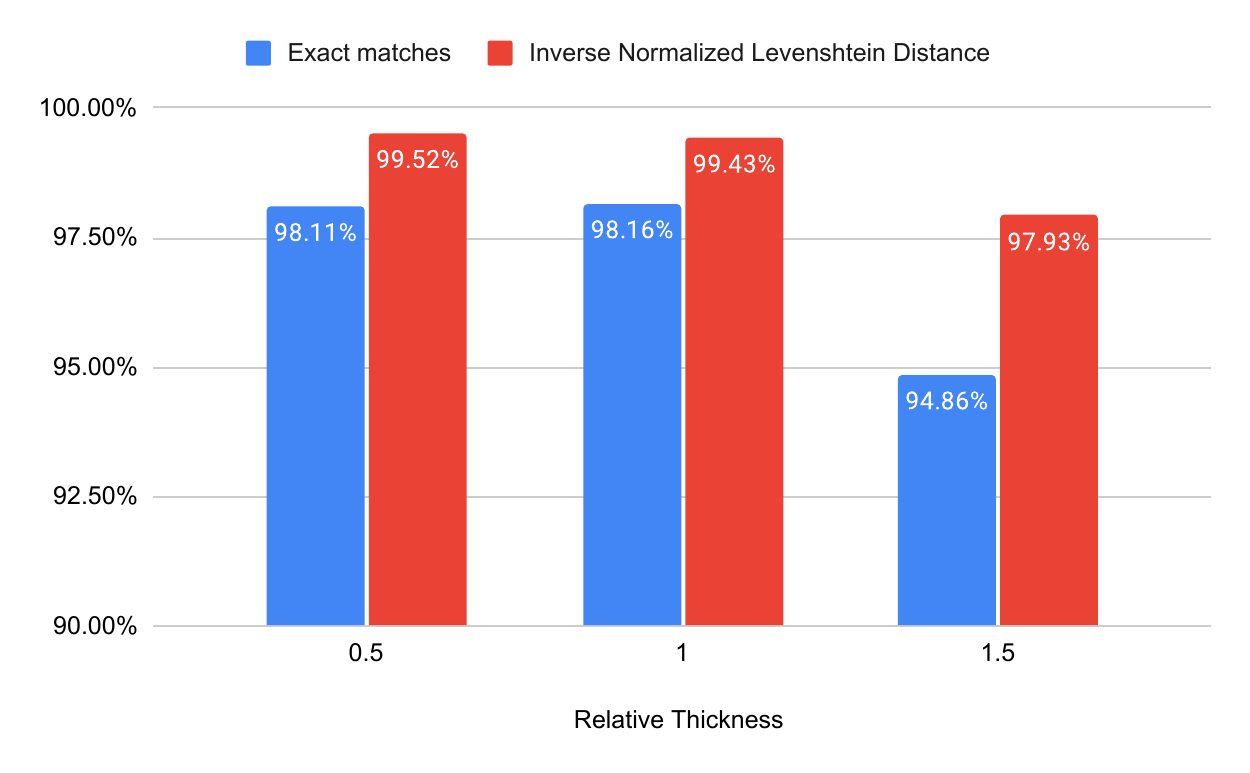}
    \caption{Sensitivity of Born-Digital Parser to Thickness Rendering Parameter. Higher thickness reduces accuracy. Other parameters: \texttt{render-implicit-hydrogens-visible} is True, \texttt{render-label-mode} is terminal-hetero.}
    \label{fig:thickness_plot}
\end{figure}

Fig. \ref{fig:label_plot} shows how different atom labelings affect  performance of the parser. Including all atom labels slightly hurts performance, in part because the more dense a molecule becomes, the more probable it is for the parser to connect atoms incorrectly.
Fig. \ref{fig:thickness_plot} then shows the effect of rendering with different thicknesses. Lower thicknesses produce stronger results, again because this decreases the density of the molecule. As seen in Fig. \ref{fig:rendering_params}, lower thickness increases the distance between unconnected objects.

\begin{figure}[!tb]
    \centering
    \includegraphics[width=0.50\textwidth]{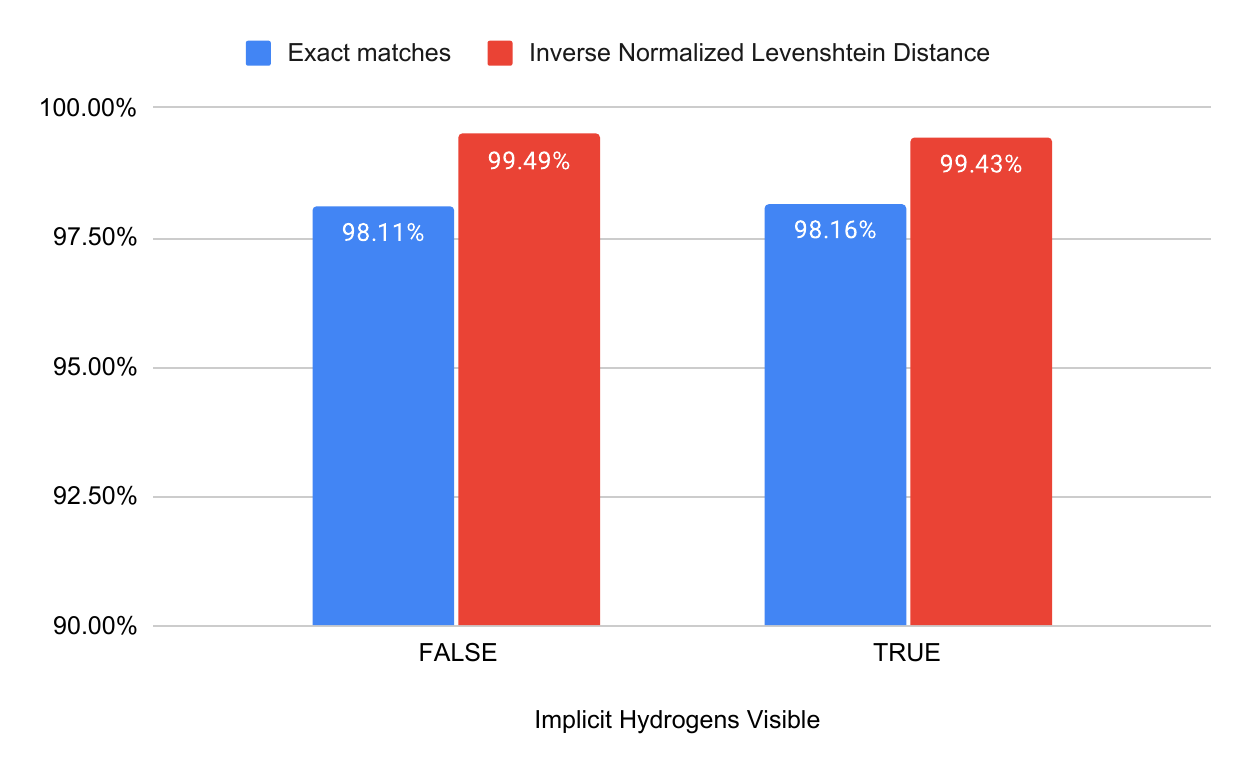}
    \caption{Sensitivty of Born-Digital Parser to Showing Implicit Hydrogens. Other parameters: \texttt{render-label-mode} is  terminal-hetero and \texttt{render-relative-thickness} is 1.}
    \label{fig:hydrogens_plot}
\end{figure}

Fig. \ref{fig:hydrogens_plot} compares performance when rendering molecules with or without implicit hydrogens. The difference between the conditions is minimal, with 14 fewer exact matches (roughly $0.06\%$) than when showing implicit hydrogens. This difference is due to merging errors of different groups that are close, similar to the crowding of Fig. \ref{fig:all_True_1.5}.

Overall, the born-digital parser is quite robust to these changes in rendering parameters. This robustness was achieved by gradually increasing the reliance of the born-digital parser on graph properties while reducing the number of  parameters used; additional reductions in parameters are likely possible.

{\bf Benchmarking: Visual Parser.}
For the synthetic USPTO dataset, our visual parser trained using outputs from our born-digital parser, obtains a recognition rate of 85.02\%. While this rate is lower than that seen for transformer-based methods like MolScribe \cite{qian_2023} and rule-based methods such as MolVec and OSRA \cite{Filippov2009}, this result still demonstrates potential. Notably, MolScribe is trained on 1.68 million examples with various chemical structure-based and image-based augmentations, and employs a SWIN transformer model with 88 million parameters. In contrast, our visual parser was trained on a much smaller dataset of 3,416 annotated images, without augmentation, and using a simpler SE-ResNeXt model with 4 million parameters. Despite these differences, our parser outperforms SWIN-OCSR \cite{xuSwinOCSREndtoendOptical2022}, which also uses a SWIN transformer but is trained on 4.5 million molecules.

We have omitted results for the real datasets (CLEF and UoB) due to limitations
in our initial training dataset, which is missing symbols from these sets and
training using a single set of Indigo rendering parameters as mentioned earlier.  This first training set does not adequately capture the diverse styles and structural variations seen in the non-synthetic data sets. We will address this in future work.
We will note here however, that the visual line primitives extracted from the real images are accurate.

We conducted training runs on the Pubchem dataset, which consisted of 
queries for 3,416 molecules in three forms: primitives, whole symbols, and symbols detected during training. Each epoch averaged 155.6 minutes, with the model completing 19 epochs in about 49 hours. This training time is notably shorter than other systems, such as DECIMER \cite{Rajan2020a}, which required 27 days to converge on 15 million structures, demonstrating efficiency with fewer data to achieve comparable results.

However, testing on the synthetic USPTO dataset (5,719 molecules) took 18.6 hours (11.74 secs/molecule), which is slower compared to systems like MolGrapher \cite{morinMolGrapherGraphbasedVisual2023} and OCMR \cite{wangOCMRComprehensiveFramework2023} that process a single molecule in less than a second. The slow inference time is due to inefficiencies in our first implementation. 
In particular, re-assembling query outputs for formulas and writing visual graphs are currently slower than they could be. Future versions will accelerate these components.

\begin{table*}[!tb]\centering
\scriptsize
\caption{
Born-Digital Parser Label Graph Metrics for Different Rendering Parameters (5719 molecules). Shown are $F_1$ measures for symbol labels, correct  labels, and complete graphs.}
\label{tab:lgeval}
\scalebox{0.75}{
\begin{tabular}{l|lll|l|l|ll}
\toprule

&\multicolumn{3}{c|}{\sc Rendering Parameters} &  \sc Correct node & \sc Correct edge&\multicolumn{2}{c}{\textbf{\sc Molecules}} \\ 
{\sc Render} &\tt label\_ mode &\tt implicit\_ hydrogens\_ visible &\tt relative\_ thickness & \sc (labels) $F_1$& \sc (labels) $F_1$& \sc Struct. & \sc +Class \\
\midrule
Default &terminal-hetero &true &1 &99.96 &99.84 &98.49 &97.62 \\


Hardest &all &true &1.5 & 99.65& 99.01 &81.89 &81.12 \\

\bottomrule
\end{tabular}
}
\end{table*}

\begin{figure*}[t!]
    \centering
    \begin{subfigure}[t]{0.5\textwidth}
        \centering
        \includegraphics[height=2.5in]{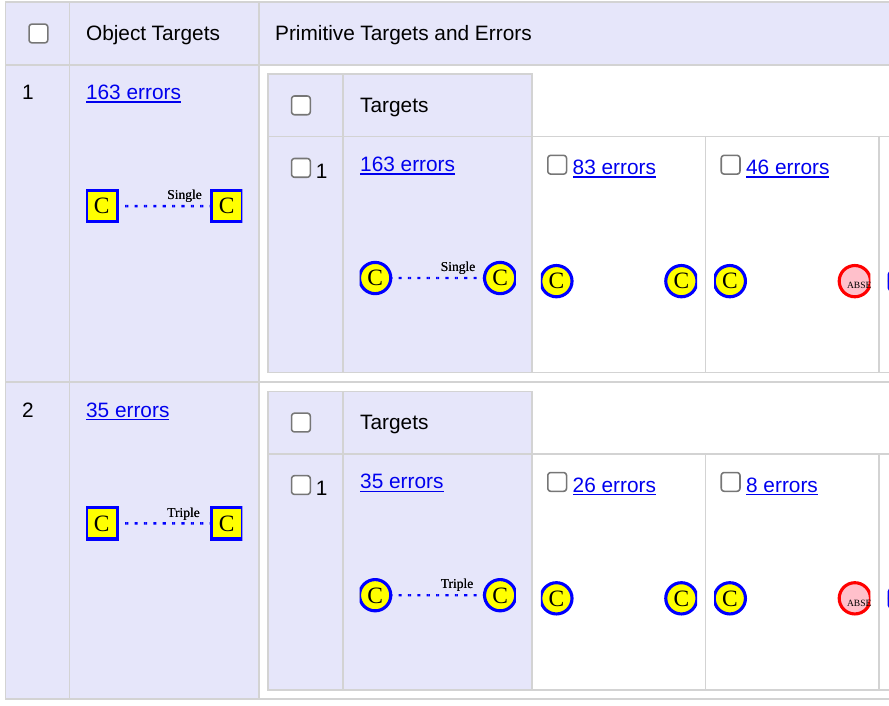}
        \caption{Default Rendering Parameters}
    \end{subfigure}%
    ~ 
    \begin{subfigure}[t]{0.5\textwidth}
        \centering
        \includegraphics[height=2.5in]{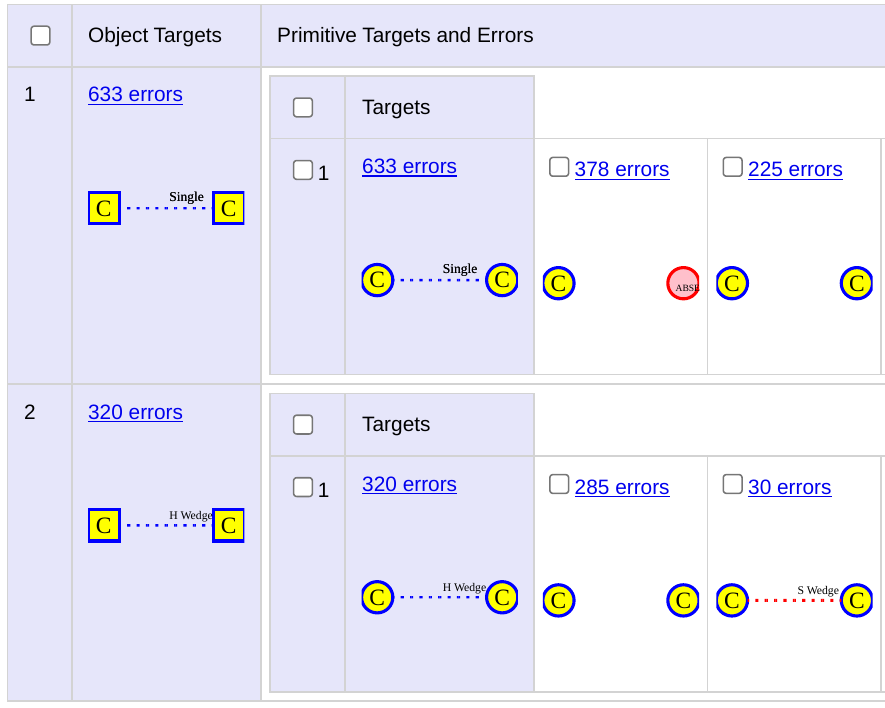}
        \caption{Hardest Rendering Parameters}
    \end{subfigure}%
    \caption{Relationship Confusion Histograms for Renderings in Table \ref{tab:lgeval} (truncated at right for space). Hyperlinks show molecules with specific errors, check boxes allow selecting molecules with errors for export. \textbf{Default rendering:} the top 2 errors are missing single and triple bonds. We can observe that in both cases, at times a missing (ABSENT) hidden carbon is the cause. \textbf{Hardest rendering:} missing single bonds are again the most frequent error, caused half of the time by a missing carbon. The second most-frequent error is missing hashed wedges between carbons, where no bond is detected, or because of misclassification of hashed wedges as solid wedges. 
    \label{fig:confHist}}
\end{figure*}

\subsection{Graph-Based Evaluation}
\label{sec:graph-eval}

For fine-grained evaluation of ChemScraper, we require
molecule graph representations for both ground truth and the predicted molecules. Given we have already created chemical structure graphs subsequently converted
to CDXML format, we can readily employ these graphs for
evaluation.
It is important to note that the molecular graphs utilized
for evaluation differ from the visual graphs created in Section \ref{sec:data} to annotate raster images.

{\bf Molecular Graphs for Evaluation.} The predicted graph corresponds to the final stage in the parsing algorithm,
shown in Fig.~\ref{fig:mst_to_final}(e). These graphs are generated in the final step of the born-digital parsing pipeline (see Fig.~\ref{fig:process}). This graph assumes the representation of atoms or atom groups as nodes,
with edges representing bond types associated with nodes, which may have one of the following types:
\texttt{\{Single, Double, Triple, Solid Wedge, Hashed Wedge\}}.
To construct a ground truth molecular structure graph, we use a \texttt{MOL} object generated by Indigo  from the corresponding SMILES representation. We then extract atom positions along with the adjacency matrix for
bonds between atoms using MolScribe code
\cite{qian_2023} with minor modifications.

We identify correspondences between nodes in parser output and ground truth graphs
using atom coordinates from Indigo (ground truth)
and Symbol Scraper (parser output). Minor discrepancies in atom
coordinates are resolved using minimum distances between
corresponding atom pairs. Corresponding nodes are giving the same identifiers.

Finally, we create object-relationship label graph files (Lg files) as described in Section \ref{sec:data}. 
`Object'
entries represent individual atoms or atom groups, and the
`Relationship' entries denote bond edges with bond type labels between the
atoms, as opposed to specifying the type of connections between visual
elements.

{\bf Analysis: Born-Digital Parser.}
We use {\tt LgEval} to compare molecular graphs to obtain the metrics in Table \ref{tab:lgeval}. The table shows a
disparity between recognition rates when using labeled graphs (last column) vs. the exact SMILES matches shown in Table \ref{tab:results}. This arises
because SMILES string-based metrics lack sensitivity to direction and errors for 3D bonds, such as hashed and solid wedge bonds. In this way, SMILES exact
matches may be misleading in terms of identifying correct molecular structures. In contrast, our graph-based metrics
readily identify such errors.

Table \ref{tab:lgeval} shows a large decline in recognition rates
when using the hardest rendering condition for the parser, despite only a 0.83\% reduction in accurate detection of edges in molecular graphs. This is mainly due to the intricate network of edges and
relationships, particularly in large structures with rings. Even a 1\% error in
relationships, as seen in the USPTO-Indigo dataset with 382,058 target relationships
for 5,719 molecules, substantially affects accuracy.

In the {\tt confHist} tool error summary (an excerpt is shown in Fig.~\ref{fig:confHist}), common errors for the default rendering include missed single and triple bonds. The run for the hardest rendering parameters produces a notable increase in the count for the most frequent errors, including missing single and hashed wedge bonds. This unexpected difficulty with easier-to-detect bonds is due to the density of molecules in the hardest rendering condition, which produces short bond lines and a compact structure (See Fig. \ref{fig:rendering_params}(b)). This poses challenges for our graph transformations using thresholds to accurately detect bonds or establish correct connections between entities. This illustrates where greater use of visual features may be beneficial within the born-digital parser itself.

{\bf Analysis: Visual Parser.}
For molecular diagrams produced by the visual parser for USPTO, symbols
including different characters, numbers, and wedges are often misclassified as
{\tt Single} bonds. This is mainly due to class imbalance in the training data
that predominantly features {\tt Single} lines (roughly 70\% of symbols in
training are single lines). Errors also include incorrect segmentations,
particularly for characters like {\tt N}, and {\tt H} that are frequently over-segmented. This is also likely due to their rarity in the training data. Additionally, relationship errors, notably missed connections between lines and characters, are comparatively more common due to the predominance of line-line connections over line-character connections.

The class imbalance in symbols and relationships, especially the predominance of the {\tt Single} class and line-line connections, highlights the need for better recognition of less frequent classes to improve the parser’s performance on diverse molecular structures. Additionally, the training set does not include all symbols present in the test sets, which impacts the parser's ability to accurately recognize and interpret a full range of molecular symbols. Addressing this imbalance and coverage is important for future enhancements.

\section{Conclusion}

We have introduced the ChemScraper born-digital molecular diagram
parser, along with improved extraction for characters and graphics from
PDF (SymbolScraper). To address a shortage of training data for molecular diagrams in raster images, we use the born-digital parser to annotate raster images with visual structure graphs.
This data is used to train a visual parser for raster images that uses a novel multi-task neural network run recurrently. 
Both the born-digital and visual parsers produce molecular structure graphs in CDXML which can be used with well-known chemical drawing tools (ChemDraw, Marvin) and easily converted to other molecular structure representations (e.g., SMILES, MOL, and InChI).

We also apply the adjacency matrix-based evaluation metrics developed for CROHME to molecular diagrams. These metrics and the LgEval tools offer
a detailed assessment of parser performance, and identify bond structure errors missing in conventional SMILES-based evaluation. 

{\bf Limitations} of this work include:

\begin{enumerate}
\item Images considered are noise-free vector and rasterized-vector images from a single rendering model (Indigo) created using a limited set of parameters. While modern PDFs contain relatively clean images, noisy images (e.g., scans of older documents) would require modified image primitives, annotation strategies, and parser designs.  
\item Born-digital parser parameters may be improved with larger grid searches, Bayesian optimization, and using visual features.
\item Graph transformations are manually defined; learned transformations may be more robust.
\item Our first visual parser has slow inference and does not yet generalize well to real images, due to limited class coverage and variation in our first training dataset.
\end{enumerate}

{\bf Opportunities} for future work include:

\begin{enumerate}
\item PDF primitives extracted by {\tt SymbolScraper} provide high-precision locations for text and graphics. This can be applied in extraction, search, and visualization applications.
\item Developing a more domain-agnostic technique for born-digital parsing. Perhaps GNNs, graph rewriting systems, or encoder-decoder models could improve results obtained from {\tt SymbolScraper} output.
\item The visual parser and graph-based evaluation methods are not domain-specific, and could be applied to other graphics including mathematical formulas and tables.
\item Applying the presented techniques to index molecules and other graphics in PDF collections for graphics-aware search applications such as MathDeck \cite{amador23}. This was the original motivation for this work, and something that we are eager to pursue.
\end{enumerate}

\subsubsection*{Acknowledgments}
This work was supported by the National Science Foundation USA (Grant \#2019897,
Molecule Maker Lab Institute). We thank Matt Langsenkamp, Matt Berry, Kate
Arneson, and other members the NCSA team who helped create the online
ChemScraper system.

\subsubsection*{Author Contributions}
S, A. K. refactored the system, enhanced some
functionalities and wrote around 35\% of the paper. D, A wrote around
20\% of the paper and coded the first version of the system. A, B wrote
around 15\% of the paper, added some functionalities, coded most of
the evaluation. C, M wrote around 20\% of the paper and refactored the
first version of parser. O, B wrote around 10\% of the paper, provided
chemist related information and feedback. D, S provided data. Z, R
helped refactor the system, lead the research, rewrote and organized the
paper.

\subsubsection*{Data Availability}
The training data generated and used in this study is
publicly available and can be accessed at 
\url{https://www.cs.rit.edu/~dprl/data/icdar2024/}. The dataset
generation script is provided in the repository code
provided in the Introduction section above.

\begin{appendices}






\end{appendices}


\bibliographystyle{splncs04}
\bibliography{refs-reduced-v2} 

\end{document}